\documentclass[sigconf,nonacm]{acmart}

\usepackage[noEnd=true]{algpseudocodex}
\usepackage{algorithm}
\usepackage{microtype}
\usepackage{multirow, multicol}
\usepackage{subcaption}
\usepackage{pgfplots}
\pgfplotsset{compat=1.8}
\usepackage{enumitem}
\setlist[itemize]{align=parleft,left=0.1em..1.2em}
\usepackage{pifont}
\newcommand{\cmark}{\ding{51}}%
\newcommand{\xmark}{\ding{55}}%
\usepackage{makecell}
\usepackage{float}
\algblock{Input}{EndInput}
\algnotext{EndInput}
\algblock{Output}{EndOutput}
\algnotext{EndOutput}
\usepackage{extdash}

\title{TimeBlocks: Foundational and Continual Time-Series Blockbase---Extended Version}

\author{David Campos}
\affiliation{%
  \institution{Aalborg University, Denmark} 
  \country{}
}
\affiliation{%
  \institution{University of Amsterdam}
  \country{Netherlands}
}
\email{dgcc@cs.aau.dk}

\author{Bin Yang}
\authornote{Corresponding authors.}
\affiliation{%
  \institution{East China Normal University}
  \country{China}
}
\email{byang@dase.ecnu.edu.cn}

\author{Tung Kieu}
\authornotemark[1]
\affiliation{%
  \institution{Aalborg University}
  \country{Denmark}
}
\email{tungkvt@cs.aau.dk}

\author{Lei Chen}
\affiliation{%
  \institution{Hong Kong University of Science and Technology, Guangzhou, China}
  \country{}
}
\email{leichen@cse.ust.hk}

\author{Chenjuan Guo}
\affiliation{%
  \institution{East China Normal University}
  \country{China}
}
\email{cjguo@dase.ecnu.edu.cn}

\author{Christian S. Jensen}
\affiliation{%
  \institution{Aalborg University}
  \country{Denmark}
}
\email{csj@cs.aau.dk}

\renewcommand{\shortauthors}{David Campos et al.}

\begin{document}

\begin{abstract}

The ongoing digitization has led to a proliferation of time-series data streams that monitor a variety of processes, from which valuable insights may be obtained.
Further, the emergence of successful foundational language models begs the question of whether it is possible to achieve time-series models with the foundational properties of handling multiple tasks, while being sufficiently lightweight to allow real-time data stream processing.
Existing foundational time-series models are often large and only effective in offline settings without stringent time and computational constraints, and where repeated model calibration is not needed.
However, when applied to data streams, these models are ineffective due to their size and lack of support for continual calibration, which compromise their ability to deliver accurate real-time responses, their durability, and their deployability in hardware-limited settings. 
We propose \texttt{TimeBlocks} to enable versatile time-series processing by facilitating the efficient building of lightweight models suitable for multiple tasks under variable conditions. In particular, the method maintains a pool of interchangeable and modular model blocks that can be used to construct new time-series models.
When presented with specific time-series data, a routing strategy iteratively selects the most suitable blocks to construct a lightweight and accurate model for the data.
We equip \texttt{TimeBlocks} with a method called \texttt{StreamCore} to build a representative small subset of the data stream, which preserves a guaranteed approximation of the stream over time, enabling continual model calibration.
An experimental study on multiple data sets and covering multiple tasks shows that \texttt{TimeBlocks} enables to build models capable of outperforming existing baselines.

\end{abstract}

\maketitle

\section{Introduction}

Due to the ongoing digitalization of societal and industrial processes, we are witnessing an increasing availability of data streams that have the potential to offer valuable insights into processes and to enable new or improved applications.
With the emergence of powerful and foundational large language models---i.e., models trained across multiple data sets and settings, thus having good generality---there is a growing interest in exploiting the impressive capabilities of such models in applications beyond language processing, such as analyzing data streams.
To enable value creation from data streams, methods must possess excellent analytical capabilities while also being lightweight and enabling real-time responses. Such methods hold the potential to enable applications in settings where decisions need to be made fast and where computational resources are limited.
For example, in intelligent vehicles, in-vehicle controllers perform real-time tasks, such as forecasting power consumption under different driving conditions~\cite{YangHGYJ23,Lu0J11} or detecting possible anomalies in the sensors~\cite{Karpathy19,GuoJ014}.
Due to hardware limitations in this type of environment, methods that are small in size and can handle multiple settings and tasks are called for.

Therefore, the problem is to develop models that can handle multiple tasks and data sets, while also being able to operate and be updated in computationally constrained environments, such as those of edge devices. This is an attractive way to handle the same type of data, i.e., time series, without deploying separate models for each task. Additionally, it enhances the capabilities of edge devices and maximizes their resource utilization. Recent studies attempt to apply foundational large language models to time series directly as these and text share sequential characteristics~\cite{ZhouZPZLXZ21,Rasul23}. However, this approach is usually inefficient, given the large sizes of language models~\cite{ZhouNW0023,WuHLZ0L23}, and it does not generalize sufficiently, as time series embody specific characteristics that benefit from more specific models~\cite{ZengCZ023,QiuHZWDZGZJSY24}.

Although a few studies aim to address some of these specific requirements of time series~\cite{Rasul23, pmlr-v235-woo24a, LiCZDNZCH22}, limitations remain that need to be addressed. In particular, existing foundational time-series models remain very large~\cite{pmlr-v235-das24c, ZerveasJPBE21,YueWDYHTX22,ZhouNW0023}.
Then, given their sizes and data requirements~\cite{ZerveasJPBE21,LiangCMIL24}, these models are not suitable in a resource-limited environment and in settings that require continual calibration or fine-tuning. This is necessary to maintain performance over time, for example, on embedded sensor controllers.
We proceed to describe two key limitations of current state-of-the-art methods for developing foundational time-series models, after which we give an overview of how to address them.

\begin{figure*}[ht!]
\begin{subfigure}{0.32\linewidth}
        \includegraphics[width=0.95\linewidth]{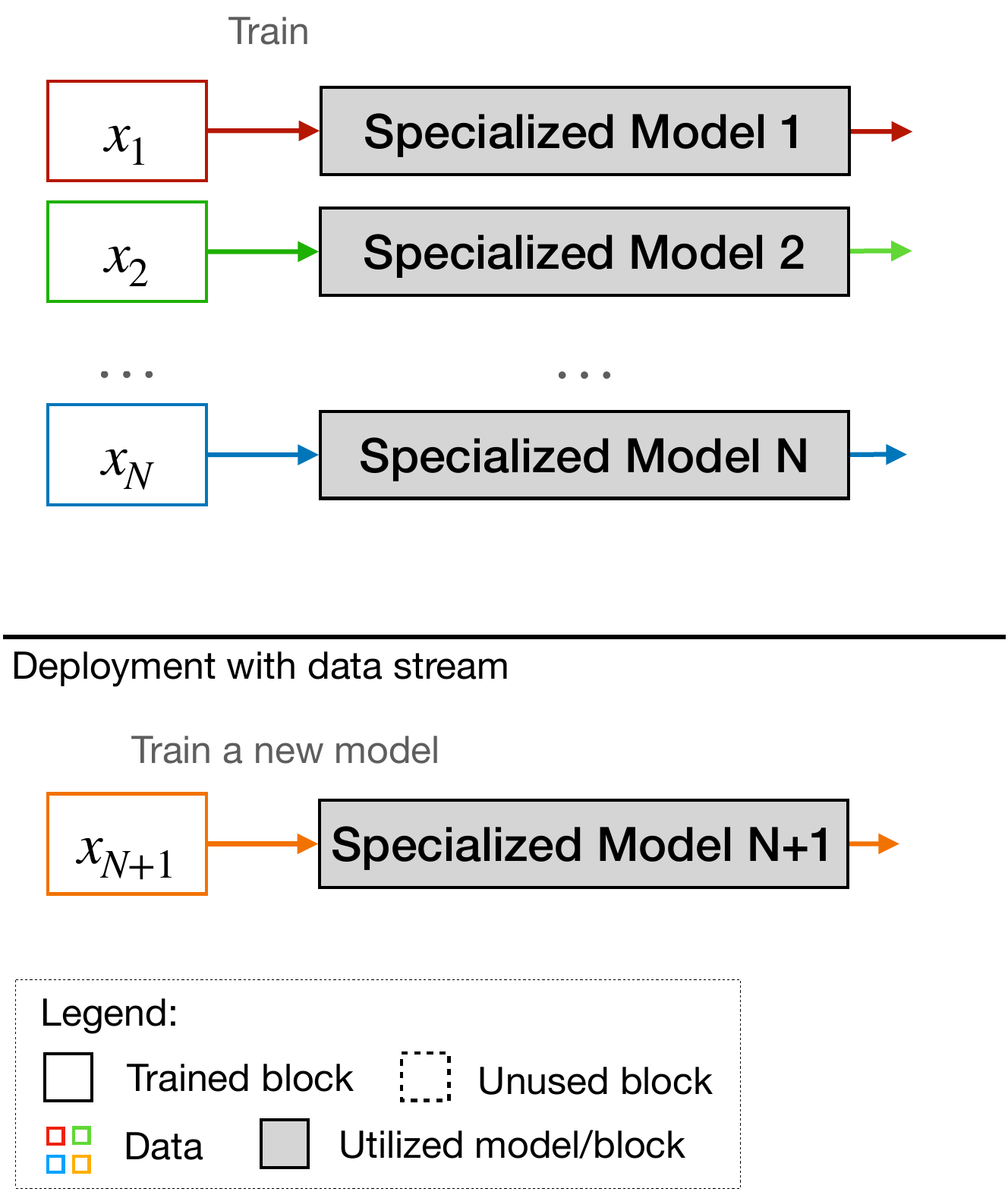}
    \caption{Specialized Models Paradigm. }
    \label{subfig:specialized}
\end{subfigure}
\begin{subfigure}{0.32\linewidth}
    \includegraphics[width=0.95\linewidth]{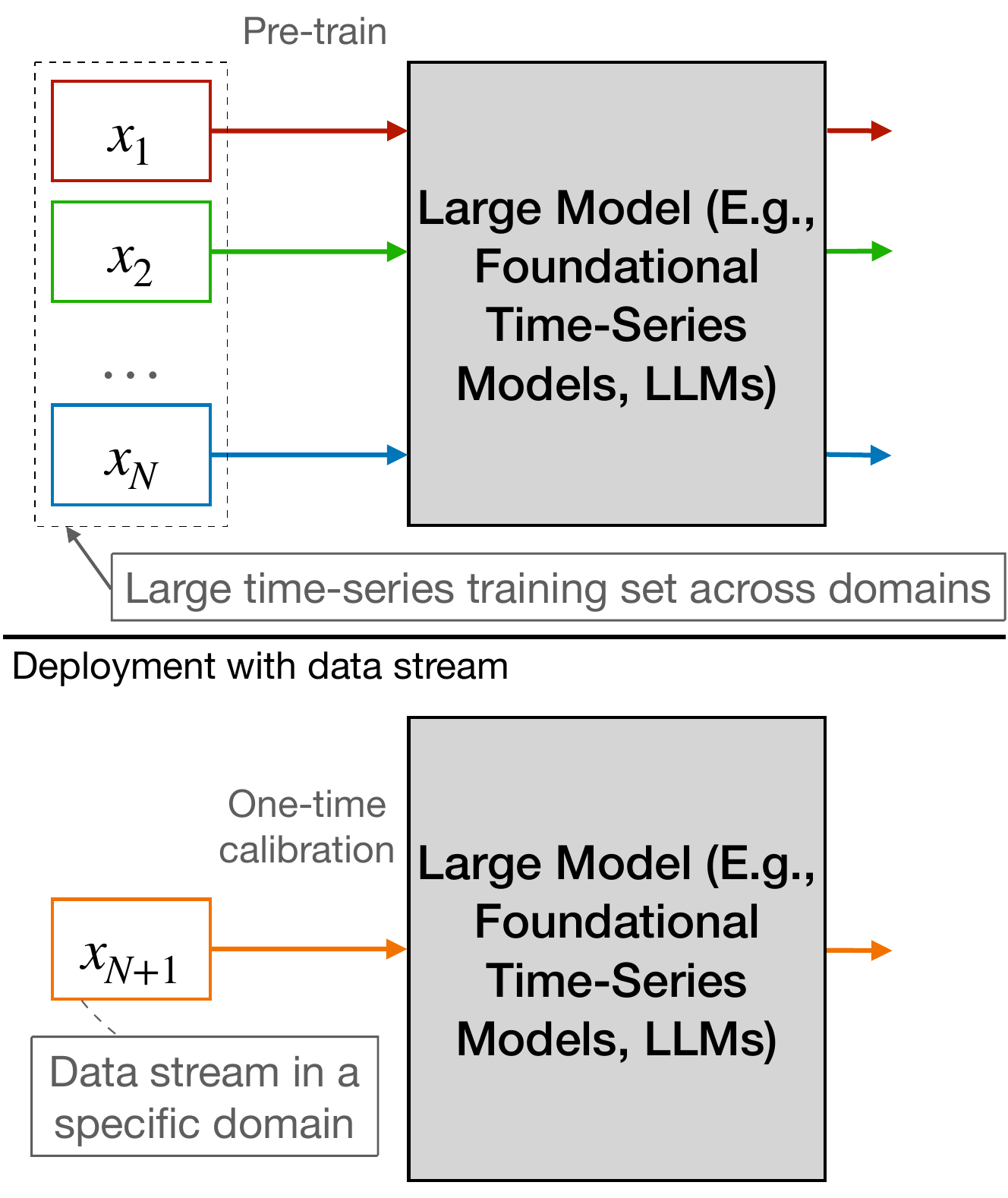}
    \caption{Large Model Paradigm.}
    \label{subfig:encoder}
\end{subfigure}
\hspace*{-3.3ex}
\begin{subfigure}{0.342\linewidth}
    \includegraphics[width=0.95\linewidth]{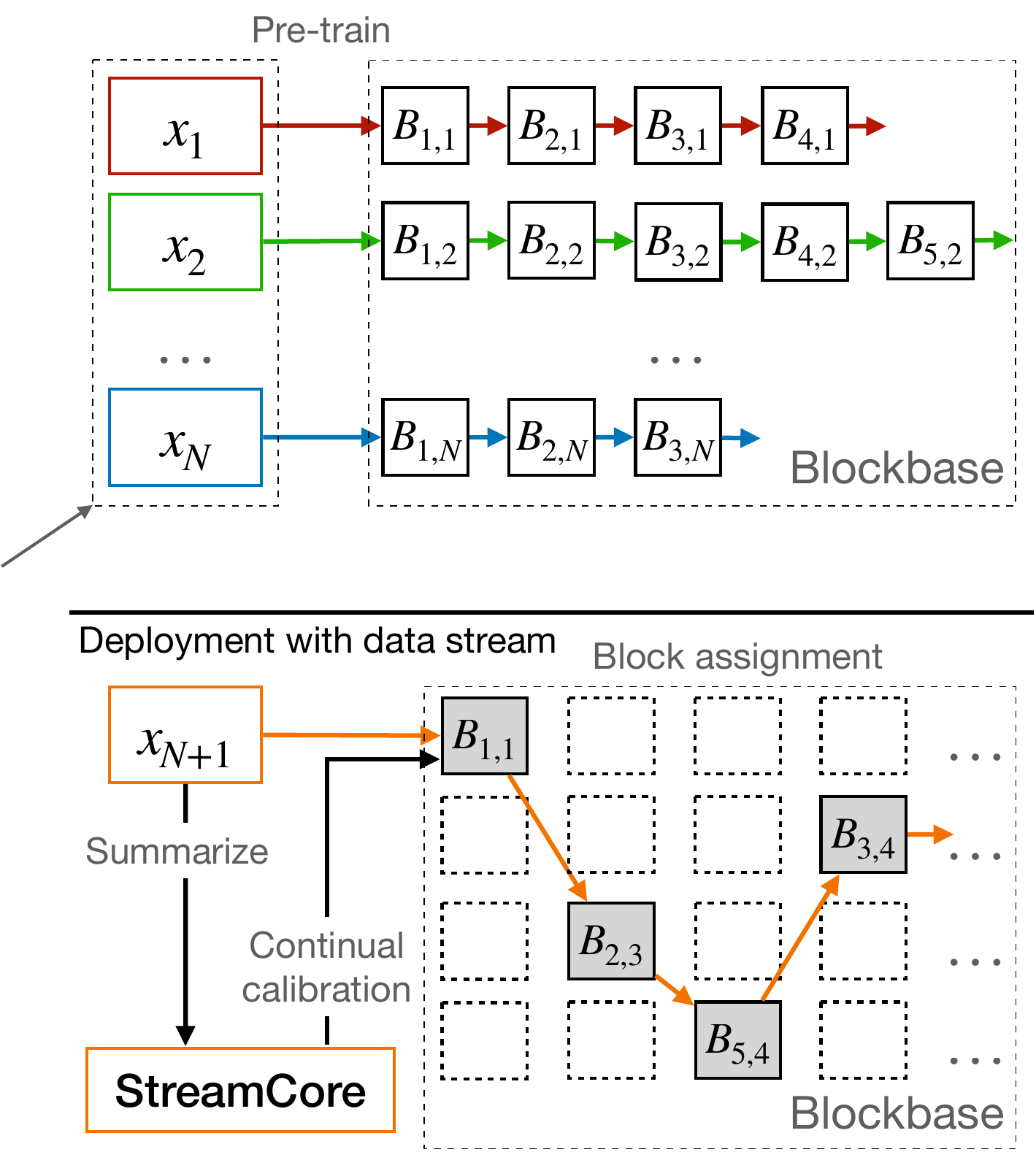}
    \caption{The \texttt{TimeBlocks} Paradigm.}
    \label{subfig:blockts}
\end{subfigure}
    \caption{Paradigms for Time-Series Processing. (a) In the specialized model paradigm, a new model must be trained for each time series. (b) In the large model paradigm, a large model is pre-trained using a large set of time series across domains $\{x_i\}_{i=1}^N$. When deployed, model calibration requires sufficient amounts of data for a single calibration step. (c) In \texttt{TimeBlocks}, multiple small time-series processing blocks are pre-trained and stored in the \texttt{Blockbase}, and a data stream is summarized into a representative subset called \texttt{StreamCore}, allowing for continual calibration.} 
    \label{fig:paradigms}
    \Description[Comparison]{Comparison}
    \vspace{-1em}
\end{figure*}

\noindent
\textbf{Large Pre-Trained Models:} 
Traditionally, when processing multiple tasks and data sets, a specialized model is trained for each case, as shown in Figure~\ref{subfig:specialized}. This approach is inefficient as it entails the creation of a substantial number of independent models. Current foundational time-series models have been developed to address this issue by allowing for the handling of multiple tasks and data sets. However, this results in larger models, as depicted in Figure~\ref{subfig:encoder}. 
This limitation suggests the need for a middle-ground solution between small, highly specific time-series models and large, more versatile foundational models, i.e., efficient small models that can handle heterogeneous time series among multiple tasks and analytic requirements.

\noindent
\textbf{Ineffective Continual Calibration:}
Given the large sizes of current foundational time-series models, calibration after deployment is usually conducted a single time due computational constraints, as shown in Figure~\ref{subfig:encoder}. This approach is suitable mostly in offline settings with sufficient computational resources to accumulate all the necessary data and perform fine-tuning or calibration.
It is not suitable in settings where the data characteristics may change over time, thus requiring continual calibration. 
Therefore, the continual calibration of large foundational models 
remains as an unresolved issue, 
as it is impractical to rapidly calibrate large models while accounting for the accumulation of sufficient data from the stream.

To address the limitations mentioned above, we propose \texttt{Time- Blocks} as a novel paradigm for continual deployment of efficient models for time-series analytics.

\noindent
\textbf{Addressing Challenge 1 (Small Model):}
Instead of using a large pre-trained model to analyze all time series, we propose a new paradigm that allows for the creation of lightweight and specialized models for time-series processing, as shown in Figure~\ref{subfig:blockts}. Rather than training a single large model, our approach involves pre-training models constituted by blocks when considering a high variety of time-series tasks, data sets, and settings, such as historical windows, or forecasting horizons. 
The notion of a block is at the core of this paradigm. A block is a tiny and independent time-series processor, that when composed with other blocks can form models to perform different types of analysis.
Thus, when designing and training models built upon blocks for time-series analytic tasks, it is possible to store the blocks of each model in a pool, that we called the \texttt{Blockbase}.
Once in the \texttt{Blockbase}, a subset of these modular time-series blocks can be connected as needed to perform different analyses on time series with different characteristics, making the approach versatile for task processing.

The proposed paradigm dynamically selects the necessary blocks at inference time when analyzing a new time series, creating a specialized model that uses only the selected blocks to analyze it. This modular approach selects the pre-trained blocks based on the performance of the model under construction, aiming to improve it with each additional block. This is different from other strategies where combining several models imply more training, such as boosting ensembles~\cite{PuruckerSABBH23} or recursive learned indexes~\cite{KraskaBCDP18}.

\noindent
\textbf{Addressing Challenge 2 (Small Data Subset):}
To enable efficient continual calibration of a lightweight model, we propose a method, called \texttt{StreamCore}. This method reduces the computational overhead of accumulating a data stream by introducing a summarization strategy. The strategy maintains a subset with minimal cost updates that align with how often the model is calibrated. Thus, \texttt{StreamCore} enables the calibration of the built model over time, while also maintaining a low computational cost to fit within constrained hardware environments.
Furthermore, the subset ensures a guaranteed representative approximation of the data stream while maintaining a low update cost.

In summary, the paper makes the following contributions:

\begin{itemize}
    \item It proposes \texttt{TimeBlocks}, a novel paradigm where independent, modular, and stackable blocks allow for the building of time-series processing models, providing a high degree of flexibility to accommodate different time series and time-series tasks.
    \item It proposes \texttt{StreamCore}, a low-cost summarization method with approximation guarantees for the continual calibration of lightweight models under time-series data streams.
    \item It reports on experiments that provide insights into the effectiveness of \texttt{TimeBlocks} for constructing models and of \texttt{StreamCore} for calibrating them under data streams.
\end{itemize}
\section{Preliminaries}

We introduce key concepts required to present the paper proposals.

\subsection{Time-Series Data Stream}

\subsubsection{Time Series}
A time series $x \in \mathbb{R}^{M \times V}$ is a time-ordered and regularly sampled sequence of $M$ timestamps, each associated with $V$ variables. When $V=1$, the time series is univariate, while when $V>1$, it is multivariate.

\subsubsection{Data Stream}
A data stream is a time series $x$ collected continuously over time, so the number of timestamps $M$ is unbounded.

\subsubsection{Analytics Function and Time Series Tasks}
Let $f(\cdot)$ be a time-series analytics function that maps a time series from $\mathbb{R}^{C \times V}$ to $\mathbb{R}^{F \times V}$, when processing a time-series task. For instance, in forecasting tasks, $f(\cdot)$ uses variables from $C$ historical timestamps to predict variables for the next $F$ timestamps, such as predicting future electricity consumption. In imputation tasks, $f(\cdot)$ maps variables from $C$ timestamps to the same space, so $F$ is equal to $C$. Imputation can be used, for instance, for reconstructing missing sensor data in cases of communication errors and for outlier detection tasks, as it is unlikely that abnormal data can be reconstructed. In classification, the time series is mapped from $\mathbb{R}^{C \times V}$ to a set $\mathbb{L}$.

\subsubsection{Context Length}
The context length is the sequence of $C$ timestamps in the domain of the analytics function $f(\cdot)$. For instance, in forecasting, the context length corresponds to the number of historical timestamps.

\subsection{Time-Series Processing Blocks} \label{ssec:processingblock}

\begin{figure}[ht]
    \centering
    \includegraphics[width=0.90\linewidth]{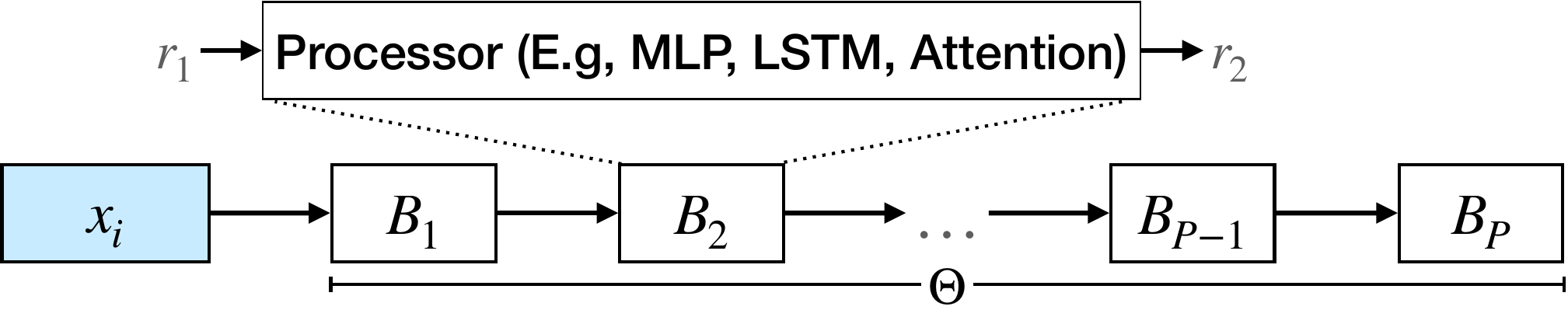}
    \caption{Time-Series Processing Blocks.}
    \label{fig:block}
    \Description[Block]{Block}
\end{figure}

Consider an analytics function $f(\cdot;\Theta)$ with parameters $\Theta$ that applies to a time series $x_i$.  
The parameters $\Theta$ may be disentangled as a sequence of $P$ processing blocks, $\Theta := B_{P}(B_{P-1}(B_{\dots}(B_{2}(B_{1}(x_i)))))$. 
Blocks may employ different architectures, including the Multilayer Perceptron (MLP), Long Short-term Memory Networks (LSTM), and attention mechanism.
The residual output $r_p$ from each block $B_{p}$ is the input to block $B_{p+1}$, as shown in Figure~\ref{fig:block}.

\section{TimeBlocks}

\subsection{Problem Setting}
We consider the problem of developing time-series processing models that can handle multiple tasks on the edge, while supporting calibration in a continual learning setting.
An example is building specialized models that enable different types of forecasting in the multiple controllers found in modern vehicles. Typically, this task is accomplished by either training new models from scratch or by fine-tuning large existing models, both of which can be time-consuming given the many models needed.
Section~\ref{ssec::block} presents the \texttt{Blockbase} routing framework that addresses these limitations. This framework enables the efficient construction of models using independent blocks that have been pre-trained. 
The method works at inference time, meaning that it builds a model 
while provides real-time results. 

Then, when a model is deployed in a continual learning setting, its operating environment may change over time due to the nature of data streams. 
Therefore, to ensure the performance of such models over time, a common technique is to perform periodic calibration with new data. However, this approach can be computationally expensive as it entails retraining of large models, despite their generality, and the accumulation of all incoming stream data.
To reduce computational costs, Section~\ref{ssec:coreset} presents means of building and maintaining a representative subset of a data stream that 
guarantees data diversity over time and enables periodic calibrations. 

\begin{figure}[ht]
    \centering
    \includegraphics[width=0.90\linewidth]{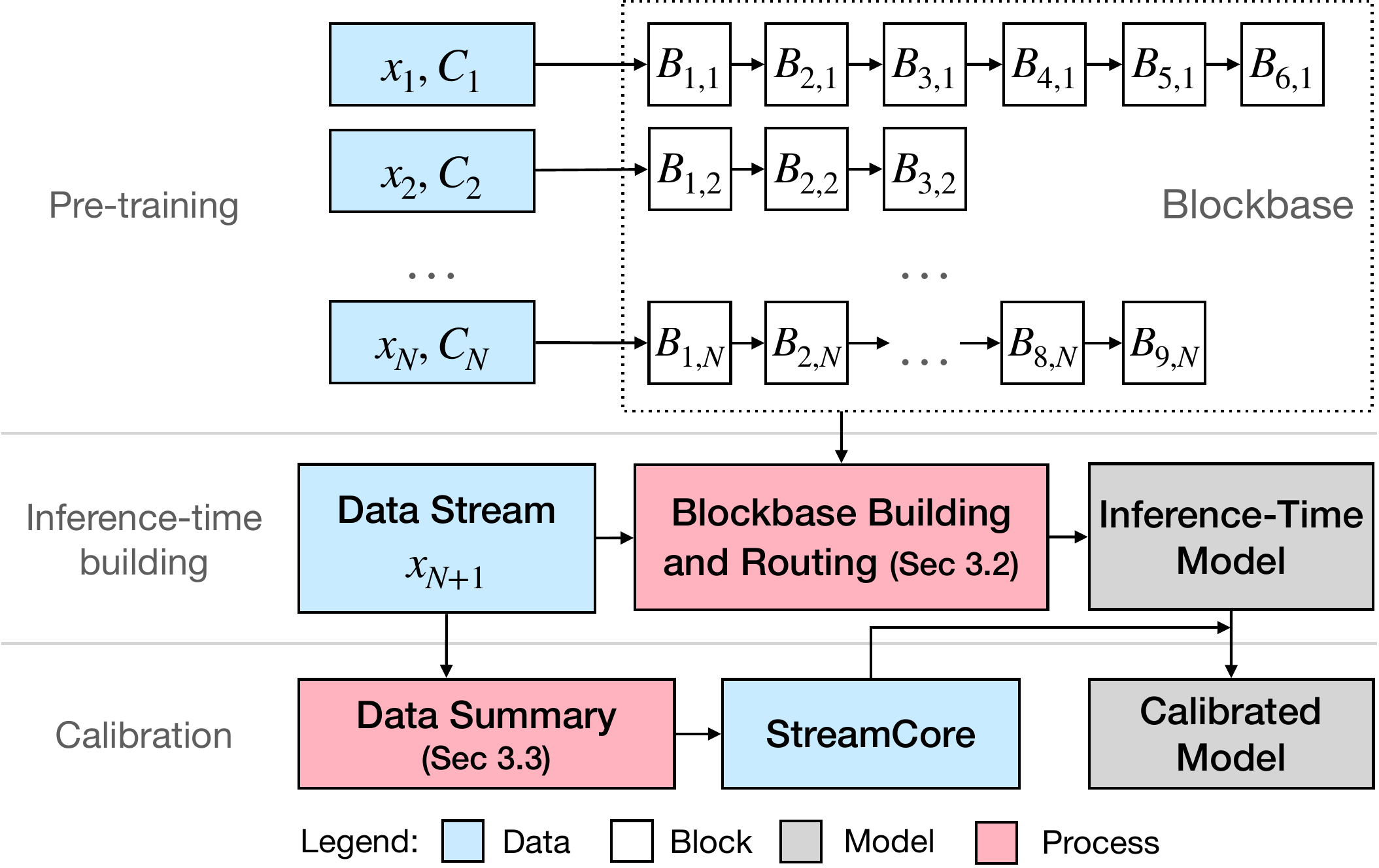}
    \caption{\texttt{TimeBlocks} Paradigm Overview.}
    \label{fig:overall}
    \Description[Overall]{Overall}
\end{figure}

An overview of the proposal is shown in Figure~\ref{fig:overall}.
First, during a pre-training stage, we generate a pool of time-series processing blocks, named \texttt{Blockbase}, by training models constituted by blocks among multiple settings. 
Second, when presented with a new data stream, a model is constructed 
by selecting and stacking the most suitable processing blocks from the \texttt{Blockbase}.
Third, to enable an efficient continual calibration, a compact representation of the data stream is maintained, so called \texttt{StreamCore}, ensuring that the most diverse data points 
are kept to facilitate continual calibration of the constructed model.

\subsection{Blockbase Routing} \label{ssec::block}

\subsubsection{Overview}

While numerous time-series models may be available for supporting multiple tasks, as shown in Figure~\ref{subfig:specialized}, the fact that they concern the same type of data suggests that 
their processing is similar. 
Existing large models, as shown in Figure~\ref{subfig:encoder}, leverage this similarity by training a one-fits-all large model to handle a wide range of time series. However, their large size limits their use in hardware-constrained settings. To address this, the proposed \texttt{Blockbase} decomposes this type of structure into independent processing blocks, as shown in Figure~\ref{subfig:blockts}.
Thus, when a new time series needs to be analyzed, a small model can be built selecting the most relevant blocks. 
The process for building and utilizing the \texttt{Blockbase} is divided into three main components, which are detailed below.

\subsubsection{Blocks Pre-Training} \label{sssec:blockTraining}

To generate blocks, we design end-to-end models using time-series processing blocks and train them using various tasks, data sets and training parameters. The complete details are available in the Appendix~\ref{appendix:block}.
Then, the trained blocks from each model are included into the \texttt{Blockbase} for future use in building inference-time models. 
An end-to-end model is, as shown in Figure~\ref{fig:block}, a sequence of independent blocks.
Each block is enhanced with a patching structure~\cite{NieNSK23} that effectively handles specific time-series conditions, such as trends and cycles~\cite{EkambaramJNSK23,abs-2401-03955}, when processing the series in smaller segments.
Also, we make the blocks modular to allow for their independent use at any position on new models when considering the following properties.

\begin{figure}[ht]
    \centering
    \vspace{-0.5em}
    \includegraphics[width=0.9\linewidth]{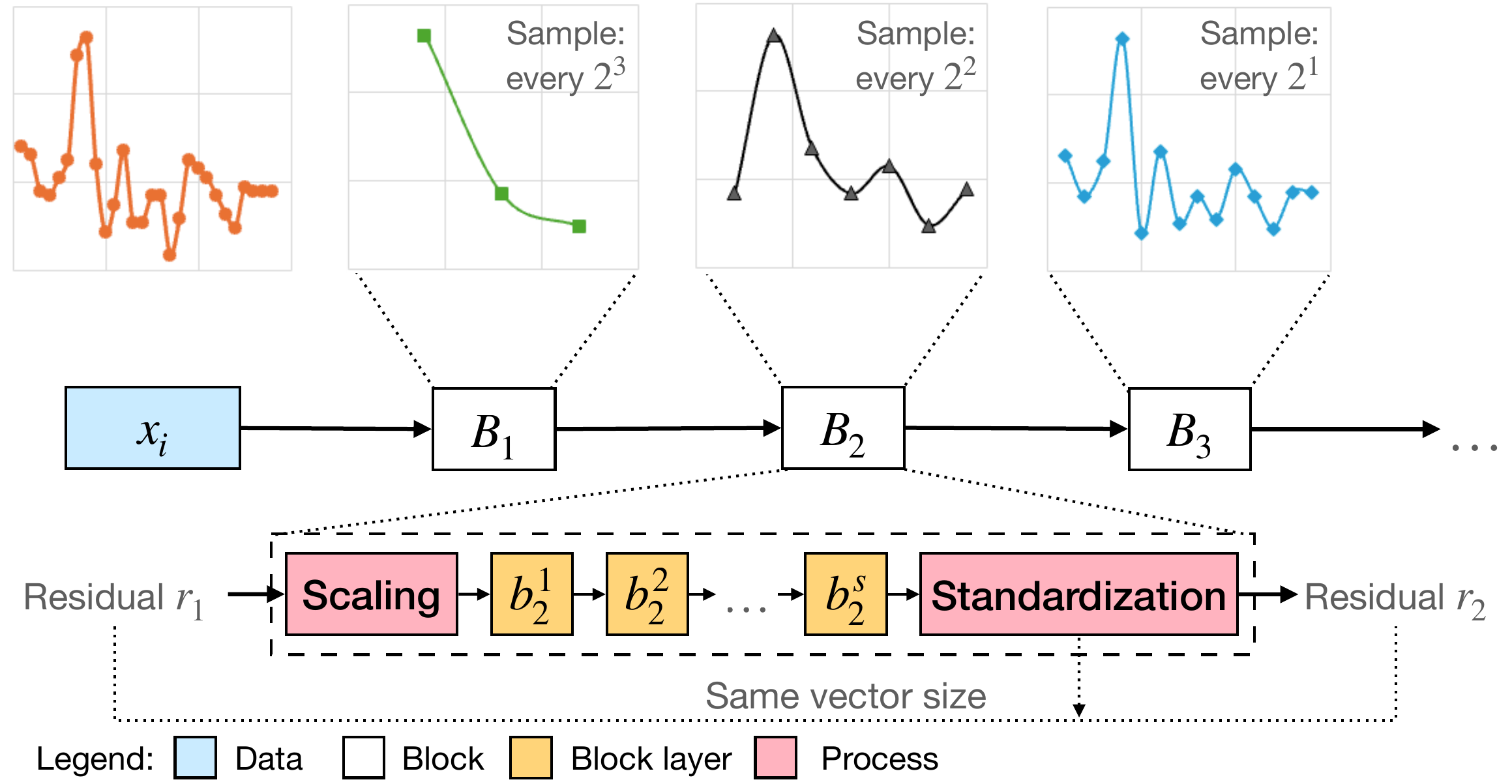}
    \vspace{-0.5em}
    \caption{Multi-Scale Block Processing.}
    \label{fig:sampling}
    \Description[Sampling]{Sampling}
    \vspace{-0.7em}
\end{figure}

\noindent
\textbf{Multi-Scale Blocks:} 
As time series exhibit high heterogeneity among frequencies~\cite{ChalluOORCD23,OreshkinCCB20}, it is important to process them while considering this variability, enabling to identify patterns at different scales.
Therefore, for a given end-to-end model, we design the blocks to process different levels of granularity depending on their sequential position within the model. Specifically, we scale the input of each block, as shown in Figure~\ref{fig:sampling} at the bottom, while following a decreasing power-of-two
pattern~\cite{ChalluOORCD23,OreshkinCCB20} to capture patterns from broad to specific. Thus, each level limits its input to double the previous level, as shown in Figure~\ref{fig:sampling}, at the top. 

In addition to enable a better processing of a time series, this multi-scale strategy ensures that the blocks included in the \texttt{Block- base} are capable of managing time series with multiple scales.

\noindent
\textbf{Output Standardization:} 
In order to achieve independence and interchangeability among every block, it is important to establish a common interface that allows the interaction and connection between any pair of blocks. 
Thus, each block $B_p$ incorporates a standardization layer for the output $r_p$, as it is shown in the below part of Figure~\ref{fig:sampling}. The operation involves a fully-connected layer that standardizes the output to a common vector size, regardless its scale or data set, enabling its compatibility with all types of blocks. 

\noindent
\textbf{Diverse Settings:}
To achieve a variety of blocks, we train end-to-end models using a wide range of data sets, tasks, architectures, training parameters, and number of blocks. For example, on forecasting tasks, we use multiple context lengths and forecasting horizons~\cite{WuQGH0G25}. 
As a result, \texttt{Blockbase} is highly diverse, the goal being to enable the processing of many types of time series when building new models.

Once all the blocks are trained, they are logically organized in the \texttt{Blockbase}, as described in Section~\ref{sssec:cluster}. Then, the router enables their selection, as will be explained in Section~\ref{sssec:router}.

\subsubsection{Blocks Clustering} \label{sssec:cluster}

As the goal of the \texttt{Blockbase} is to have a large pool of time-series block processors, it is important to design a strategy to logically organize them efficiently in order to ensure a speedy block retrieval process. 

Consider a block $B_p = \langle b_p^1, b_p^2, \dots, b_p^s \rangle$ with $s$ layers (e.g., a MLP with $s$ linear layers). After training the block, the weights of all its layers remain constant, allowing these weights to serve as a means of distinguishing one block from another. In particular, we designate the first layer, $b_p^1$, as the primary identifier for the block, since this is the layer where the block first receives data flow from a previous block. We referred to it as the \textit{fingerprint} because it serves as a distinctive identifier for the block training. Thus, the \textit{fingerprint} $b_p^1 \in \mathbb{R}^{\mathit{In} \times V \times \mathit{Out}}$ is a homogeneous weight vector across all blocks, where $\mathit{In}$ and $\mathit{Out}$ represent the input and output sizes of the layer, while $V$ refers to the number of variables.

To organize logically the blocks after their training, we cluster all the generated blocks using their fingerprint as the metric of comparison. 
Clustering provides flexibility as it allows for some fingerprints to be assigned to the most suitable cluster instead of directly assigning a group based solely on their parameters.
The clustering assignment is done using the $K$-Means algorithm~\cite{Jin2010} with $K$ equal to the number of considered context lengths, as experiments show that blocks trained with the same context length sometimes, but not always, share similar fingerprints.
$K$-Means is applied to the complete set of blocks, and employs the block that is closest to the cluster center for further fingerprint comparisons.

\subsubsection{Router} \label{sssec:router}

To build a new model using blocks in the \texttt{Blockbase}, 
we propose a routing mechanism that, when given a time series, iteratively builds a model by identifying the blocks that are most beneficial for processing it. 

\begin{figure}[ht]
    \centering
    \vspace{-0.5em}
    \includegraphics[width=0.75\linewidth]{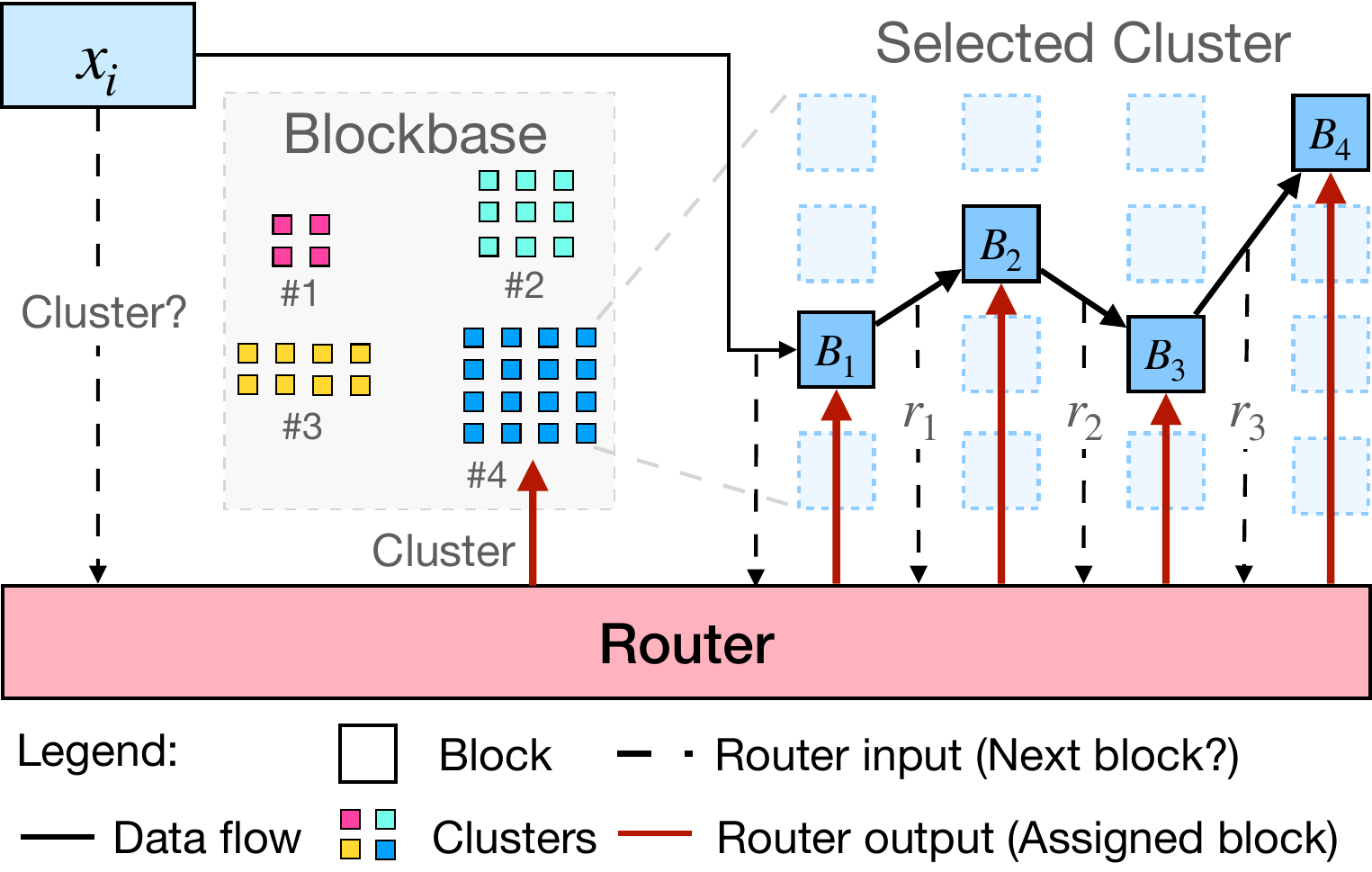}
    \vspace{-1em}
    \caption{Router Overview.}
    \label{fig:router}
    \Description[Router]{Router}
    \vspace{-1.5em}
\end{figure}

An overview of the router functionality is shown in Figure~\ref{fig:router}. Initially, when presented with a time series, the router identifies the most appropriate cluster of blocks. From there, it can select the most suitable block within the cluster, thereby also determining the context length needed for analyzing the time series. This is determined by the first block as it is the only one that processes the input time series directly.
Then, the router selects each block based on the result of evaluating the model up to the last added block. For example, it selects the second block when evaluating the model using only one block. The selected blocks may include blocks with different scales, but there is no restriction on following a pattern. This process repeats until the required number of blocks is reached.

To design the routing mechanism, we use model features to assess the model status during the construction process. With the model parameters denoted as $\Theta := B_{P}(B_{P-1}(B_{\dots}))$, the residual $r_p$ of each block $B_p$ transitions to block $B_{p+1}$, as shown in Figure~\ref{fig:block}. This residual serves as a means to assess the model until block $B_p$.
To map the relationship between the block residual $r_p$ and the following block $B_{p+1}$, we employ the block fingerprint $b_{p+1}^1$ (see Section~\ref{sssec:cluster}) as an identifier of the block, since it represents the first part of the block that interacts with its input.
Therefore, we define a router function that establishes the link between two consecutive blocks as $Router(r_p) \approx b_{p+1}^1$. This router aims to approximate the fingerprint of a block that matches the output $r_p$. Furthermore, each block is designed to produce a standardized output (see Section~\ref{sssec:blockTraining}), ensuring that the mapping remains consistent across all residuals.

\noindent
\textbf{Block Selection Problem:}
%
The router is a heuristic proposed to efficiently identify a set of blocks, a process that is otherwise computationally expensive. 
Thus, the problem of selecting blocks for an inference-time model can be viewed as choosing a path of length $j$ within the \texttt{Blockbase} that yields optimal performance.
This problem can be considered a derivation of the Hamiltonian path problem, as detailed below.

Consider a set of blocks $\mathcal{B} = \{B_1, B_2, \dots, B_N\}$ as the vertices of a graph. A path between two blocks $\mathcal{B}^\prime \subset \mathcal{B}$ has a length given by the number of edges between them, and a cost given by the weight defined for those edges.  
For a given time-series task, it is reasonable to assume that a Hamiltonian path can be built, since the modular design of the blocks allows the $N$ vertices to be connected by a path of length $N-1$.  
Finding such a path is an NP-hard problem, while, if such a path exists, there is one with minimal cost.  
Thus, the problem can be reduced to the case $j=N-1$, which is defined by the number of blocks requested to build a model.

To evaluate the cost of a path $c(\cdot)$, the simplest approach is to measure the performance gain from adding each block. However, this is prohibitively expensive, as it requires evaluating all blocks for every addition. Therefore, we must use a different type of metric to select the path.

To address this problem, alongside approximating an optimal solution for the Hamiltonian path, we introduce a router as a heuristic to enable the selection of blocks based on how the models are structured when trained independently. Thus, the metric that defines the cost between two blocks is implicitly learned when training the router. This training minimizes a loss between interconnected blocks, enabling the router to identify blocks that are similar to those that have processed analogous data in the independent models.

Considering that the router can approximate a Minimum Spanning Tree, as it is possible connects all vertices while minimizing total cost, and that every vertex can be associated with an edge (i.e., the graph has a minimal perfect matching), it has been shown under these assumptions~\cite{Hoogeveen91} that the approximation for a Hamiltonian path $\mathcal{B}^\prime_{ij}$ between two vertices $B_i$ and $B_j$ has a tight bound of $5/3$ relative to the optimal solution $\mathcal{B}^*_{ij}$. Therefore, the cost of the path found by the router is bounded as follows:
\begin{align}
    c(\mathcal{B}^\prime_{ij}) \leq c(Router(\mathcal{B}^\prime_{ij})) \leq \frac{5}{3} c(\mathcal{B}^*_{ij}) \nonumber
\end{align}

\noindent
\textbf{Router Training:} To obtain a $Router(\cdot)$ that can link consecutive blocks, we train an auxiliary model to learn to do so during the block pre-training stage. 
At this stage, the blocks of each model are optimally interconnected, and the data flows between the blocks is similar to what would be required when building a model in inference time.
The architecture of the router is kept simple, consisting of two 1D convolutional layers followed by a fully-connected layer, as it is important to be able to select blocks efficiently.

\begin{figure}[ht]
    \centering
    \vspace{-0.5em}
    \includegraphics[width=1\linewidth]{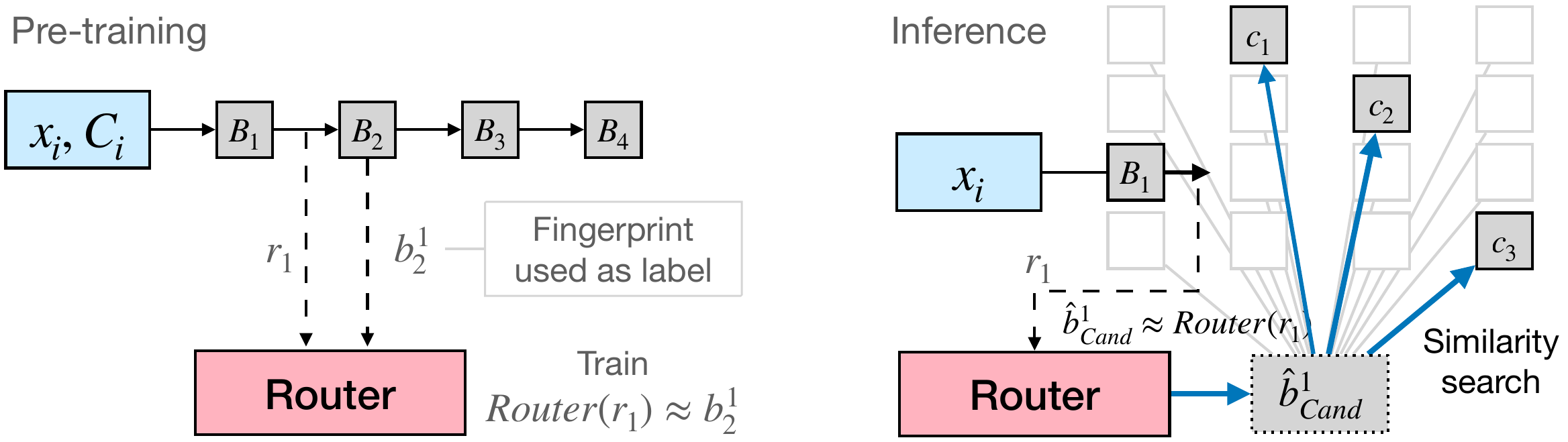}
    \vspace{-1em}
    \caption{Router Training and Block Selection.}
    \label{fig:blockSelection}
    \Description[BlockSelection]{BlockSelection}
    \vspace{-1em}
\end{figure}

The process for training $Router(\cdot)$ is shown in Figure~\ref{fig:blockSelection} to the left. 
In this example, the router is trained to approximate the fingerprint $b_{2}^1$ of the next block $B_2$ using the residual $r_1$ from block $B_1$, such that $\text{Router}(r_1) \approx b_{2}^1$. The router is then trained across all blocks of the model and other trained models. The completed details are available in the Appendix~\ref{appendix:router}.

\subsubsection{Inference-Time Model Building} \label{sssec:zeroshot}

To efficiently build an inference\Hyphdash time model, we use the \texttt{Blockbase} in conjunction with the router to select the blocks that are most suitable for processing a given time series. The router first selects the optimal cluster for processing the time series, thereby reducing the number of blocks to be considered for model construction.

\noindent
\textbf{Context Length Selection:}
Noticing that when selecting a block, we are also choosing its context length, we use this property to evaluate different context lengths when searching for a block and then choose the most suitable one. 
Further, as we organize the blocks into clusters to enhance search efficiency, as described in Section~\ref{sssec:cluster}, the choice of context length is determined by evaluating potential clusters. This is because, in general, we experimentally observe that blocks within a cluster tend to share the same context length.  
Thus, as Figure~\ref{fig:router} shows, when given a time series, the router, using the \texttt{Blockbase}, identifies the closest cluster while also determining the context length. 
This reduces the number of evaluations needed to select new blocks, another router task to be detailed shortly, as the selection of blocks is done limited to the cluster. Overall, the comparison uses the minimum distance as the metric for selecting blocks, as it indicates how closely the fingerprints are aligned and is fast to compute. More complex metrics could also be considered, provided they remain simple and effective. 

\noindent
\textbf{Block Selection:}
After selecting the cluster from the \texttt{Blockbase}, the router builds the model by identifying the most suitable blocks within it.
Considering a given time series $x$, the objective is to identify a suitable subset of blocks that allows to process $x$ such as the analytics function (e.g., forecasting) has minimal error, while staying within a predetermined budget of $J$ blocks.
Formally, when considering the \texttt{Blockbase} as a set $\mathcal{B}$, the goal is to identify a set $\mathcal{B}^\prime \subset \mathcal{B}$, $\mathcal{B}^\prime = \{B_1, B_2, \dots, B_J\}$, for the time series $x$ and the analytics function $f(\cdot)$ so that when evaluating $f(\cdot)$ using $\mathcal{B}^\prime$ yields the lowest possible error, while using at most $J$ blocks, i.e., $|\mathcal{B}^\prime| \leq J$.
To process a time series, the blocks in the set $\mathcal{B}^\prime$ are connected sequentially. Each block takes an input, processes it, and directs the output residual to the next block for further processing. Therefore, the objective is to minimize a criterion $g(\cdot,\cdot)$ that aligns subsequent pairs of blocks to form a directed path $B_1 \rightarrow B_2 \rightarrow \dots \rightarrow B_J$, which can be formulated as follows.
\begin{align}
    \arg \min_{\mathcal{B}^\prime \subset \mathcal{B}} \sum_{j = 1}^{\mathcal{B}} g(B_j,B_{j+1}) \quad s.t. \; |\mathcal{B}^\prime| < J \nonumber
\end{align}

To select subset $\mathcal{B}^\prime$, the router is used in an iterative process, as illustrated in Figure ~\ref{fig:blockSelection} to the right, until reaching $J$ blocks.
Each time a block $B_j$ is selected, the model is evaluated up to that block, and the resulting residual $r_j$ is sent to the router to enable it to select the next block $B_{j+1}$. In Figure~\ref{fig:blockSelection} to the right, the output $r_1$ from block $B_1$ is used by the router to calculate the candidate fingerprint for the second block $\hat{b}_{\mathit{Cand}}^1$. Then, through similarity search, the router compares this fingerprint with those in the cluster and selects the blocks with the most similar fingerprints, denoted as $c_1,c_2,c_3$, which can be performed with a cost on the order of $J \times \log(|\mathcal{B}|)$~\cite{JohnsonDJ21}. 
The complete details are available in the Appendix~\ref{appendix:router_eval}.

As the search process identifies a group of candidate blocks, multiple independent paths can be evaluated simultaneously, 
while ensuring that the results are not dependent solely on a single subset $\mathcal{B}^\prime$. 
Thus, when assessing multiple tasks like forecasting and outlier detection, 
we can determine the most appropriate subset for each specific task, instead of applying a single subset to all tasks. 

The selection of blocks is accelerated using vector processing techniques~\cite{abs-2405-13954} and it is conducted at inference time. This results in a model tailored to the data stream in which it is used. We also enable periodic model calibration to accommodate any shifts in the data stream, as time series data can change due to variable environments or seasonality. This process is explained next.

\subsection{StreamCore} \label{ssec:coreset}

When considering how to calibrate a model with a data stream, one important issue is how to effectively accumulate the data. Simple solutions, such as logging all the data, can be costly in terms of storage and computation, while basic buffer may not accurately represent the stream. Therefore, we consider computing a subset that can approximate the data stream while maintaining a small size to effectively calibrate the model and limiting the possible information loss. To achieve this, we first need to prove that building such a subset is feasible and efficient, as detailed as follows.

Consider an analytics function $f(\cdot)$ and a data stream $\mathcal{D}$, which are evaluated by a loss function $l(\cdot,\cdot)$. A subset $\mathcal{S} \subset \mathcal{D}$ is represented by a non-negative vector $w = [w_1,\dots,w_n]$, where 
$w_i > 0$ indicates that a time series $x_i \in \mathcal{S}$. Then, a subset $\mathcal{S}$ of size $m$ is considered a coreset of $\mathcal{D}$ when
\begin{gather}
    w^* \in \arg\min_{w, \lVert w \rVert_0 = m} L(\mathcal{D},w) \nonumber
\end{gather}
\begin{gather}
    L(\mathcal{D},w) = \frac{1}{n}  \Bigg\lVert \sum^n_{i\in \mathcal{D}} l({x}_i,f(x_i)) - \sum^n_{j\in \mathcal{S}} w_j l({x}_j,f(x_j)) \Bigg\rVert \label{Eq2}
\end{gather}
achieves highly similar performance with respect to evaluating the data set $\mathcal{D}$.
Then, considering a streaming setting, a new data set $\mathcal{D}^t$ arrives at time $t$, so the goal becomes finding a subset $\mathcal{S}$ such that Equation~\ref{Eq2} becomes
\begin{gather}
    w^* \in \arg\min_{w^t, \lVert w^t \rVert_0 = m} L(\mathcal{D}^t,w^t), \label{Eq3}
\end{gather}
where the subset $\mathcal{S}$ is updated at every $t$ to maintain examples from the complete feature space.

Solving the coreset selection is usually an intractable problem since it is necessary to evaluate the function for all the examples~\cite{MaaloufEMFO24,TukanBFR21}. Also, for streaming settings, the computation should be efficient, so it becomes necessary to provide guaranteed approximations.

By considering Lipschitz continuity, Equation~\ref{Eq2} can be bounded as follows (the derivation is provided in the Appendix~\ref{appendix:streamcore}):
\begin{gather}
    \frac{1}{n} \Bigg\lVert \sum^n_{i\in \mathcal{D}} l({x}_i,f(x_i)) - \sum^n_{j\in \mathcal{S}} w_j l({x}_j,f(x_j)) \Bigg\rVert \leq \frac{1}{n} \sum^n_{i\in \mathcal{D}} \min_{j \in \mathcal{S}} \lambda \lVert x_i - x_j \rVert \nonumber
\end{gather}

Thus, an approximate solution for the coreset can be found depending exclusively on the data without computing the function $f$ for all the examples, i.e., the boundary. Therefore, it is only necessary to calculate the distance in the feature space of the examples to identify a subset that minimize the overall distances. This corresponds to the facility location problem, an NP-hard problem~\cite{MegiddoS84}, so the goal is to provide solutions with approximation guarantees.

There are some solutions to this problem~\cite{LiSC22,MirzasoleimanBL20} that provide guarantees to the offline method. They are typically $\alpha$-approximate algorithms~\cite{Meyerson01}, meaning that their results are within a factor $\alpha$ from the optimal solution. For instance, $\alpha=2$ algorithms exist with a running time of $\Phi = \mathcal{O}(|\mathcal{S}|\times|\mathcal{D}^t|)$.

A simple extension of these algorithms to the streaming setting~\cite{LiSC22} involves computing the coreset and merging it every time $t$. The typical cost of each of these operations is $\mathcal{O}(|\mathcal{S}| \times |\mathcal{D}^t|)$ over $T$ periods, resulting in a total running time of $\mathcal{O}((|\mathcal{S}| \times |\mathcal{D}^t|)^2 \times T)$.
As this is computationally expensive, we introduce \texttt{StreamCore} as a method to reduce the cost of building subsets. 
It achieves this by avoiding the merging operation when updating a coreset. 

Considering an existing $\alpha$-approximate solution for the facility location, it has been demonstrated that it is possible to maintain a $(8\alpha + 4)$-approximation for the same problem by at most $R = \frac{\Phi}{4\alpha}$ arbitrary updates with at most a linear cost~\cite{Cohen-AddadHPSS19}. Thus, consider a simple algorithm that adds every new point with cost $\mathcal{O}(1)$ until the size of the coreset is met. Once the size has been reached, for each new point added, an existing point should be removed. Randomly selecting an existing point in $\mathcal{S}$ and deleting the point that is closer has a cost of at most  $|\mathcal{S}|$ while maintains the coreset diverse, as removals are uniformly distributed.

Then, as each time $t$ the data stream includes $|\mathcal{D}^t|$ observations, the period is reduced to $R = \frac{\Phi}{4\alpha \times |\mathcal{D}^t|}$ as there are $|\mathcal{D}^t|$ updates every $t$. Thus, for example, a \texttt{StreamCore} of size $|\mathcal{S}| = 400$ can be updated after receiving data over 50 periods when computing $R$.

The solution maintains a total cost of $\mathcal{O}(|\mathcal{S}| \times |\mathcal{D}^t| \times T)$, unlike the squared cost required by having to merge operation of multiple subsets, while ensuring a constant and guaranteed 20-approximation, as a linear cost update exists for an $2$-approximation algorithm. 
To perform calibration using the \texttt{StreamCore} $\mathcal{S}$ after $R$ steps, the built small model is first frozen for all blocks except the last one and then calibrated using $\mathcal{S}$. This condition prevents catastrophic forgetting and maintains model performance. 
\section{Experiments}

\subsection{Experimental Setup}
\subsubsection{Tasks and Data Sets}

We consider four different tasks: forecasting, imputation, outlier detection and classification. For forecasting and imputations tasks we use popular forecasting data sets~\cite{ZhouZPZLXZ21}, namely \textit{ETTh1}, \textit{ETTh2}, \textit{ETTm1}, and \textit{ETTm2}, which measure Electricity Transformer Temperature at different frequencies, and \textit{Weather}, for meteorological data. For outlier detection, we use the \textit{SMD} data set~\cite{SuZNLSP19} for web server monitoring, \textit{MSL}~\cite{HundmanCLCS18} for spatial exploration, \textit{SMAP}~\cite{HundmanCLCS18} for soil moisture, and \textit{SWaT}~\cite{MathurT16} for water treatment. We evaluate classification using the UCR time-series archive~\cite{DauBKYZGRK19}.

\subsubsection{Metrics}

To evaluate performance on forecasting and imputations tasks, we consider 
Mean Squared Error (\textit{MSE}). 
For the outlier detection task, we use the F1 Score metric, with a self-determined threshold~\cite{GoswamiCCMK23}. For classification, we evaluate and rank \textit{Accuracy}. 

\subsubsection{Baselines}

To evaluate the effectiveness of the the proposal, we compare it with the following baselines from three different categories. First, as specialized time-series models, we consider \texttt{Autoformer}~\cite{WuXWL21}, \texttt{TimesNet}~\cite{WuHLZ0L23}, \texttt{PatchTST}~\cite{NieNSK23}, \texttt{DCdetector}~\cite{YangZZW023}, \texttt{AnomalyTransformer}~\cite{XuWWL22}, \texttt{InceptionTime}~\cite{FawazLFPSWWIMP20}, and \texttt{ResNet}~\cite{WangYO17}. As foundational time-series models, we include \texttt{TimeMixer}~\cite{WangWSHLMZ024}, \texttt{Moment}~\cite{GoswamiSCCLD24}, \texttt{SymTime}~\cite{wang2025synthetic},
and \texttt{Tiny Time Mixers (TTM)}~\cite{abs-2401-03955}. And finally, \texttt{Lag-Llama}~\cite{Rasul23}, as a LLM-based model. 

\subsection{Experimental Results}

For the following experiments, we present the average results, while detailed results are available in the Appendix~\ref{appendix:experiments}.

\subsubsection{Inference-Time Model for Forecasting}
To compare our proposal with other baselines, we first evaluate the inference-time model performance, without any fine-tuning or calibration. The setting follows common practices~\cite{abs-2401-03955,WuHLZ0L23}, where each model, the baselines and our proposal, is trained on a large group of data sets and then tested on one unseen data set. 
For example, when we pre-train on the data sets \textit{ETTh1}, \textit{ETTh2}, \textit{ETTm1}, and \textit{ETTm2}, we then only test on the unseen \textit{Weather} data set.

When comparing the average inference-time model performance (across 96, 192, 336, and 720 horizons) in Table~\ref{table:baselines_zero}, \texttt{TimeBlocks} achieves the best results with a substantial margin over all the baselines. 
In most cases, the differences show statistical significance for $t$-tests. The large improvements over the large pre-trained models is evidence of the flexibility of selecting the most suitable blocks, rather than relying on a one-fits-all approach that cannot accommodate inference time evaluation and automatic context length selection. 

\begin{table}[ht!]
\addtolength{\tabcolsep}{-1pt}
    \small
    \centering
    \caption{Forecasting Inference-Time Model (Average MSE).}
    \label{table:baselines_zero}
    \begin{tabular}{ |l|*{6}{c|} } 
    \hline
    \textbf{Model} 
    & \textit{ETTh1} & \textit{ETTh2} & \textit{ETTm1} & \textit{ETTm2} & \textit{Weather} \\
    \hline
\multirow{1}{*}{\texttt{Autoformer}} 
 & 1.009 & 0.587 & 0.823 & 0.502 & 0.353 \\ 
\multirow{1}{*}{\texttt{PatchTST}} 
 & 0.675 & 0.385 & 0.629 & 0.293 & 0.250 \\ 
\multirow{1}{*}{\texttt{TimeMixer}}
 & 0.716 & 0.373 & 0.585 & 0.302 & 0.238 \\ 
\multirow{1}{*}{\texttt{TimesNet}} 
 & 0.792 & 0.416 & 0.738 & 0.330 & 0.257 \\ 
\multirow{1}{*}{\texttt{Lag-Llama}} 
 & 0.750 & 0.419 & 0.660 & 0.345 & 0.260 \\ 
\multirow{1}{*}{\texttt{Moment}} 
 & 0.512 & 0.655 & 0.729 & 0.710 & 0.732 \\ 
\multirow{1}{*}{\texttt{SymTime}}
 & 0.414 & 0.365 & 0.356 & 0.265 & 0.234 \\ 
\multirow{1}{*}{\texttt{TTM}} 
 & 0.428 & 0.363 & 0.545 & 0.302 & 0.236 \\ 
\multirow{1}{*}{\texttt{TimeBlocks}} 
 & \textbf{0.416} & \textbf{0.338} & \textbf{0.344} & \textbf{0.242} & \textbf{0.220} \\ 
 \hline
\end{tabular}
\end{table}

\subsubsection{Continual Forecasting}

In a continual learning scenario, where the inference-time model is deployed in a data stream, we consider a setting in which data arrives continuously and needs to be accumulated in a buffer for periodic fine-tuning. To establish baselines, we accumulate a random 5\% of the incoming data, as this is common when calibrating static models~\cite{ZhouNW0023}. Similarly, the proposed \texttt{StreamCore} also retains only 5\% of the data. For consistency, we calibrate all models at the same number of steps $R=10$, which is below the result presented in Section~\ref{ssec:coreset}, to enable several low-cost calibrations.

When comparing the performance of the model after calibrating it in a data stream, as shown in Table~\ref{table:baselines_continual}, \texttt{TimeBlocks} consistently achieves the best average results across all settings. 
The results also demonstrate a notable improvement in the baselines, compared to the inference-time model setting shown in Table~\ref{table:baselines_zero}. This highlights the strong dependence of these models on the data on which they were trained. Therefore, their performance is affected markedly when used in a continual learning setting when new data is arriving. 
In contrast, the improvements in \texttt{TTM} and \texttt{TimeBlocks} after calibration are relatively restrained. This suggests that these models perform well in the inference-time mode already, with \texttt{TimeBlocks} achieving a much more substantial performance improvement.

\begin{table}[ht!]
\addtolength{\tabcolsep}{-1pt}
    \small
    \centering
    \caption{Forecasting Continual Model (Average MSE).}
    \label{table:baselines_continual}
    \begin{tabular}{ |l|*{6}{c|} } 
    \hline
    \textbf{Model} 
    & \textit{ETTh1} & \textit{ETTh2} & \textit{ETTm1} & \textit{ETTm2} & \textit{Weather} \\
    \hline
\multirow{1}{*}{\texttt{Autoformer}} 
 & 0.529 & 0.520 & 0.587 & 0.334 & 0.336 \\ 
\multirow{1}{*}{\texttt{PatchTST}} 
& 0.464 & 0.385 & 0.374 & 0.280 & 0.240 \\ 
\multirow{1}{*}{\texttt{TimeMixer}} 
& 0.461 & 0.387 & 0.381 & 0.278 & 0.233 \\ 
\multirow{1}{*}{\texttt{TimesNet}}  
& 0.457 & 0.404 & 0.406 & 0.289 & 0.256 \\ 
\multirow{1}{*}{\texttt{Lag-Llama}} 
 & 0.459 & 0.398 & 0.420 & 0.283 & 0.248 \\ 
\multirow{1}{*}{\texttt{Moment}} 
 & 0.462 & 0.602 & 0.718 & 0.701 & 0.723 \\ 
\multirow{1}{*}{\texttt{TTM}} 
 & 0.429 & 0.361 & 0.440 & 0.280 & 0.232 \\ 
   \hline
\texttt{TimeBlocks + Random} 
& 0.404 & 0.337 & 0.348 & 0.249 & 0.225 \\  
\multirow{1}{*}{\texttt{TimeBlocks}} 
 & \textbf{0.393} & \textbf{0.334} & \textbf{0.342} & \textbf{0.244} & \textbf{0.220} \\ 
 \hline
\end{tabular}
\end{table}

\subsubsection{Coreset Types}

In order to evaluate the performance of the coreset strategy, we compared it with random selection in the last two rows of Table~\ref{table:baselines_continual}. The results show that the performance is consistently better for \texttt{StreamCore}, as it is able to maintain a better subset for the calibration. The random strategy does not provide a tight guarantee~\cite{Phillips16}, so there may be cases where performance does not improve due to a sub-optimal selection. 
We do not evaluate other strategies since they rely primarily on computing gradients~\cite{BuzzegaBPAC20,ChaudhryRRE19}, which is not applicable to an inference-time built model.

\subsubsection{Inference-Time Model for Imputation}

To assess the flexibility of our proposal, we study its performance on imputation tasks, where a portion of the data set is missing and the goal is to reconstruct the input. In this evaluation, we use an inference-time model setting, using the same set of blocks, 
and to ensure a fair comparison of the baselines, we have trained these models on imputation tasks. 
The average results among three three levels of missing data (10\%, 20\%, and 30\%) are shown in Table~\ref{table:baselines_imputation}. \texttt{TimeBlocks} demonstrates a consistently strong performance across the data sets. This highlights its flexibility in handling tasks other than forecasting, especially considering that the baselines were specifically trained for imputation tasks.

\begin{table}[ht!]
    \small
    \vspace{-1em}
    \centering
    \caption{Imputation Inference-Time Model (Average \textit{MSE}).}
    \label{table:baselines_imputation}
    \begin{tabular}{ |l|*{5}{c|} } 
    \hline

\textbf{Model} & \textit{ETTh1} & \textit{ETTh2} & \textit{ETTm1} & \textit{ETTm2} & \textit{Weather} \\
 \hline 
\texttt{Autoformer} & 0.890 & 0.468 & 0.630 & 0.395 & 0.365 \\
\texttt{PatchTST} & 0.567 & 0.334 & 0.421 & 0.308 & 0.281 \\
\texttt{TimeMixer} & 0.715 & 0.348 & 0.440 & 0.301 & 0.269 \\
\texttt{TimesNet} & 0.650 & 0.354 & 0.486 & 0.326 & 0.286 \\
\texttt{Lag-Llama} & 0.627 & 0.338 & 0.440 & 0.300 & 0.266 \\
\texttt{Moment} & 0.759 & 0.404 & 0.551 & 0.363 & 0.330 \\
\texttt{TTM} & 0.527 & 0.300 & 0.394 & 0.251 & 0.245 \\
\texttt{TimeBlocks} & \textbf{0.518} & \textbf{0.298} & \textbf{0.385} & \textbf{0.248} & \textbf{0.234} \\ 
 \hline
\end{tabular}
\vspace{-1em}
\end{table}

\subsubsection{Inference-Time Model for Outlier Detection}
\label{inference-outlier}

The results for the inference-time outlier detection model are shown in Table~\ref{table:outliers}. \texttt{TimeBlocks} achieves better performance than most of the baselines. This highlights the flexibility of the method in handling tasks other than forecasting and imputation, a capability that is lacking in the baselines, such as the otherwise competitive \texttt{TTM}.

\begin{table}[ht!]
    \small
    \centering
    \caption{Outlier Detection Inference-Time Model (F1 Score).}
    \label{table:outliers}
    \begin{tabular}{ |l|*{4}{c|} }
    \hline
\textbf{Model} & \textit{SMD} & \textit{MSL} & \textit{SMAP} & \textit{SWaT} \\ 
\hline 
\texttt{Autoformer} & 0.746 & 0.740 & 0.809 & 0.827 \\ 
\texttt{PatchTST} & 0.823 & 0.818 & 0.819 & 0.822 \\ 
\texttt{TimeMixer} & 0.738 & 0.744 & 0.809 & 0.901 \\ 
\texttt{TimesNet} & 0.813 & \underline{0.861} & 0.864 & 0.879 \\ 
\texttt{Lag-Llama} & 0.789 & 0.785 & 0.767 & 0.767 \\ 
\texttt{AnomalyTran} & \underline{0.836} & 0.851 & \underline{0.865} & 0.894 \\ 
\texttt{DCdetector} & 0.807 & \textbf{0.875} & 0.862 & \underline{0.914} \\ 
\texttt{Moment} & 0.749 & 0.727 & 0.735 & 0.815 \\ 
\texttt{TTM} & 0.783 & 0.787 & 0.762 & 0.883 \\
\texttt{TimeBlocks} & \textbf{0.854} & 0.848 & \textbf{0.866} & \textbf{0.915} \\ 
 \hline 
\end{tabular}
\end{table}

\subsubsection{Inference-Time Model for Classification} \label{inference-classification}

\begin{figure}[ht]
    \centering
    \vspace{-0.5em}
    \includegraphics[width=1\linewidth]{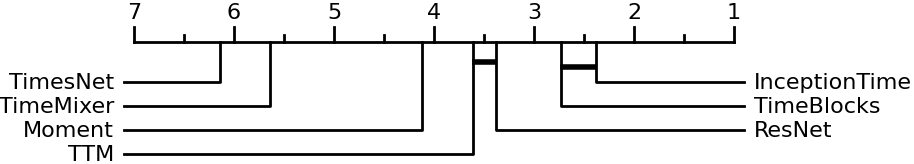}
    \vspace{-1em}
    \caption{Classification Ranking (Accuracy).}
    \label{fig:classification}
    \Description[Classification]{Classification}
    \vspace{-1.5em}
\end{figure}

The results for classification using an inference-time model are shown in Figure~\ref{fig:classification}, including only the baselines capable of handling this type of task. The figure, which uses a critical difference rank, shows that \texttt{TimeBlocks} achieves equivalent \textit{Accuracy} performance to the task-specific baseline on the complete data set.

\subsubsection{Model Size}

\begin{figure}[ht]
\small
\centering
\vspace{-0.5em}
\begin{subfigure}{0.5\linewidth}
\begin{tabular}{ |l|r| } 
    \hline
    \textbf{Model} &  \textit{Size (MB)} \\
    \hline
\texttt{Autoformer} & 2449.604 \\
\texttt{PatchTST} & 1920.776 \\
\texttt{TimeMixer} & 25.476 \\
\texttt{TimesNet} & 241.680 \\
\texttt{Lag-Llama} & 576.528 \\
\texttt{Moment} & 151.641 \\
\texttt{AnomalyTran} & 29.472 \\
\texttt{TTM} & 3.176 \\
\texttt{TimeBlocks $J=6$} & 2.985 \\
\texttt{TimeBlocks $J=2$} & 0.995 \\
    \hline
    \end{tabular}

    \caption{All Baselines.}
    \label{subfig:modelSize}
\end{subfigure}
\hspace*{-1.5ex}
\begin{subfigure}{0.5\linewidth}
    \begin{tikzpicture}
        \begin{axis}[
            xlabel=Size (MB),
            ylabel=MSE,
            ymin=0.261,
            ymax=0.30,
            xmax=3500,
            xmin=250,
            xtick = {1000,2000,3000},
            xticklabels={1,2,3},
            width=1.05*\linewidth,
            height=0.54*\axisdefaultheight,
            legend style={
                    at={(-0.1,-0.4)},
                    anchor=north west,
                    legend columns=2,},font=\footnotesize]
            \addplot[blue,mark=*] table[x=Size, y=TimeBlocks] {Figures/Data/TTM.txt} [yshift=8pt,font=\tiny]
            node [pos=0]    {$J=1$}
            node [pos=0.2,yshift=2pt]  {$J=2$}
            node [pos=0.4]  {$J=3$}
            node [pos=0.6]  {$J=4$}
            node [pos=0.8,yshift=-15pt]  {$J=5$}
            node [pos=1]    {$J=6$};
            \addlegendentry{\texttt{TimeBlocks}}
            \addlegendimage{black,dashed,mark=*}
            \addplot [color=black, dashed] table[row sep=crcr]{0   0.285213\\ 3600   0.285213\\};
            \addlegendentry{\texttt{TTM}}
            \addplot[black,mark=*] table[x=Size, y=TTM] {Figures/Data/TTM.txt} [yshift=8pt,font=\tiny]
            node [pos=1]    {$J=6$};
        \end{axis}
    \end{tikzpicture}
    \caption{Number of Blocks, \textit{ETTh2}.}
    \label{subfig:TTMSize}
\end{subfigure}
\caption{Model Size.}
\label{fig:sizeComparison}
\Description[Efficiency]{Efficiency}
\vspace{-0.5em}
\end{figure}
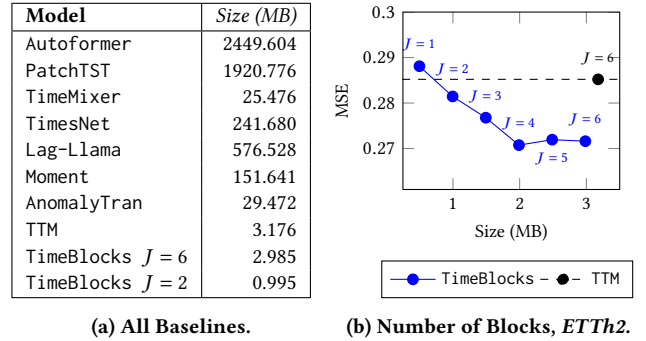

When comparing the average model size among data sets, see Figure~\ref{subfig:modelSize}, \texttt{TimeBlocks} is the smallest model, making it suitable for deployment in environments with hardware limitations. This result is expected given the design strategy focusing on building a model with only the necessary blocks for a given setting, instead of using general large models. 
The improvement in terms of size with respect to \texttt{TTM} is slight due to their use of similar small architectures. However, \texttt{TTM} has a fixed design with six blocks, while \texttt{TimeBlocks} allows for the use of fewer blocks than this default, effectively reducing its size while maintaining better performance than \texttt{TTM}, as shown in Figure~\ref{subfig:TTMSize}. 
Thus, it is possible for \texttt{TimeBlocks} to achieve better performance than \texttt{TTM}
by using $J=2$ blocks with one third of the \texttt{TTM} size, as shown in Figure~\ref{subfig:modelSize} when $J=2$. 

\subsubsection{Model Efficiency} 

\begin{figure}[ht]
\small
\centering
\vspace{-1em}
\begin{subfigure}{0.5\linewidth}
    \begin{tikzpicture}

\begin{axis}[
    width=1*\linewidth,
    height=0.5*\axisdefaultheight,
	xlabel=Model,
    ylabel=Time (s) per prediction,
    ylabel style={align=center,text width=2cm},
    xtick={1,2,3,4,5,6,7,8},
    xticklabels={Autoformer,PatchTST,TimeMixer,TimesNet,Lag-Llama,Moment,TTM,TimeBlocks},
    x tick label style={font=\tiny, rotate=40, xshift=-15pt,yshift=0pt,anchor=north},
    xtick style={style={draw=none}},
]

\addplot[black,fill,fill opacity=0.2,ybar,bar width=5pt] coordinates {(5,1.755277778)};
\addplot[blue,fill,fill opacity=0.2,ybar,bar width=5pt] coordinates {(1,1.17939083)};
\addplot[red,fill,fill opacity=0.2,ybar,bar width=5pt] coordinates {(2,1.039921615)};
\addplot[gray,fill,fill opacity=0.2,ybar,bar width=5pt] coordinates {(3,0.553226944)};
\addplot[teal,fill,fill opacity=0.2,ybar,bar width=5pt] coordinates {(4,0.387951023)};
\addplot[brown,fill,fill opacity=0.2,ybar,bar width=5pt] coordinates {(7,0.403111111)};
\addplot[purple,fill,fill opacity=0.2,ybar,bar width=5pt] coordinates {(8,0.286525)};
\addplot[olive,fill,fill opacity=0.2,ybar,bar width=5pt] coordinates {(6,0.515323161)};

\end{axis}

    \end{tikzpicture}
    \caption{Inference Time, \textit{ETTh2}.}
    \label{subfig:runtime}
\end{subfigure}
\hspace*{-1.5ex}
\begin{subfigure}{0.5\linewidth}
    \begin{tikzpicture}
        \begin{axis}[
            xlabel=Percentage,
            ylabel=MSE,
            ymin=0.261,
            ymax=0.30,
            xtick = {1,3,5,10,20},
            width=1*\linewidth,
            height=0.5*\axisdefaultheight,
            legend style={
                    at={(-0.1,-0.4)},
                    anchor=north west,
                    legend columns=2,},font=\footnotesize]
            \addplot[red,mark=*] table[x=Block, y=ETTh2] {Figures/Data/Coreset.txt};
            \addlegendentry{\textit{ETTh2}}
            \addplot[black,mark=*] table[x=Block, y=ETTm1] {Figures/Data/Coreset.txt};
            \addlegendentry{\textit{ETTm1}}
            \addplot [color=black, dashed] table[row sep=crcr]{5   0\\ 5   0.5\\};
        \end{axis}
    \end{tikzpicture}
    \caption{\texttt{StreamCore} Size.}
    \label{subfig:core_size}
\end{subfigure}
\caption{Model Efficiency. Forecasting horizon is 96.}
\label{fig:efficiency}
\Description[Efficiency]{Efficiency}
\vspace{-1em}
\end{figure}
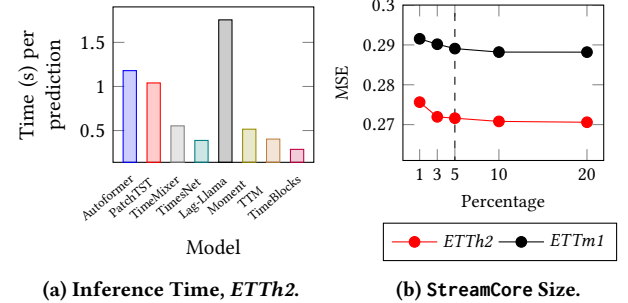

When evaluating the running time in Figure~\ref{subfig:runtime}, \texttt{TimeBlocks} achieves the lowest inference time, which is due mainly to the small size of the model. This demonstrates the advantage over the baselines, which are bigger and slower, even when the blocks for the model need to be selected. 

Then, in Figure~\ref{subfig:core_size}, we show the effectiveness of the \texttt{Stream- Core} for continual calibration across different subset sizes. 
%
The results show that using a subset size of more than 5\% does not improve performance. Therefore, it becomes a balanced choice for continual calibration, as it clearly improves the performance of smaller subsets.

\begin{figure}[ht]
\small
\centering
\vspace{-0.5em}
\begin{subfigure}{0.5\linewidth}
    \begin{tikzpicture}
        \begin{axis}[
            xlabel=Blockbase size ($|\mathcal{B}|$),
            ylabel=Time (s),
            width=1*\linewidth,
            height=0.5*\axisdefaultheight,
            legend style={
                    at={(-0.03,-0.45)}, 
                    anchor=north west,
                    legend columns=2,},font=\footnotesize]
            \addplot[red,mark=triangle] table[x=Blocks, y=Six] {Figures/Data/BlocksSearch.txt};
            \addlegendentry{$J=6$}
            \addplot[blue,mark=diamond] table[x=Blocks, y=Two] {Figures/Data/BlocksSearch.txt};
            \addlegendentry{$J=2$}
        \end{axis}
    \end{tikzpicture}
    \caption{Scalability, \textit{ETTh2}.}
    \label{subfig:search}
\end{subfigure}
\hspace*{-3ex}
\begin{subfigure}{0.5\linewidth}
    \begin{tikzpicture}
        \begin{axis}[
            xlabel=Number of blocks ($J$),
            ylabel=Time (s),
            width=1*\linewidth,
            height=0.5*\axisdefaultheight,
            legend style={
                    at={(-0.25,-0.45)}, 
                    anchor=north west,
                    legend columns=2,},font=\footnotesize]
            \addplot[black,mark=triangle*] table[x=Blocks, y=Regular] {Figures/Data/SearchTime.txt};
            \addlegendentry{Full-search}
            \addplot[blue,mark=diamond*] table[x=Blocks, y=TimeBlocks] {Figures/Data/SearchTime.txt};
            \addlegendentry{\texttt{TimeBlocks}}
        \end{axis}
    \end{tikzpicture}
    \caption{Building Time, \textit{ETTh2}.}
    
    \label{subfig:searchTime}
\end{subfigure}
\vspace{-0.5em}
\caption{Model Efficiency. Forecasting horizon is 96.}
\label{fig:efficiency_appendix}
\Description[Efficiency]{Efficiency}
\vspace{-1.5em}
\end{figure}
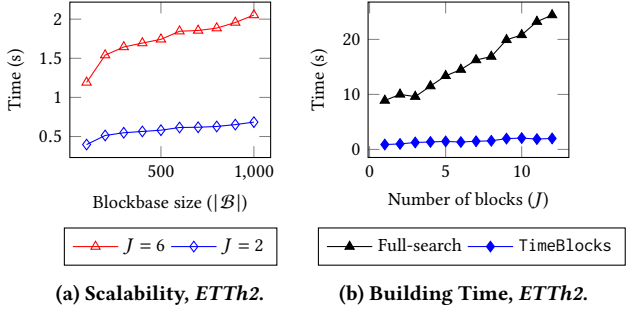

When evaluating the building time of models of two and six blocks, while adjusting the number of available blocks in the \texttt{Block-} \texttt{base}, i.e., varying $|\mathcal{B}|$, the scalability of the search process is logarithmic, as shown in Figure~\ref{subfig:search}. The search efficiency is achieved by the use of blocks with fingerprints and a router that facilitates their fast retrieval from the \texttt{Blockbase}, as detailed in Section~\ref{sssec:zeroshot}.

The building time of \texttt{TimeBlocks} models is evaluated in Figure~\ref{subfig:searchTime} when varying the number of possible blocks in a model. It is compared with a simple full-search strategy that does not use a router; it is necessary to evaluate all the actual blocks. The results show that the router is effective at maintaining a low execution time with a relatively small growth rate, as discussed in Section~\ref{sssec:router}.

\section{Related Work}

We discuss the related studies focusing on three main aspects: foundational time-series processing, streaming subset building, and model ensembling.

\noindent
\textbf{Foundational Time-Series Processing:} 
Recently, there has been a growing interest in building foundational time-series models to handle multiple data sets and settings, in an attempt to replicate the success of large language models~\cite{abs-2302-13971} when processing text. 
We review relevant studies along two dimensions: the support of continual calibration and the model size, as summarized in Table~\ref{table:relatedwork}.

\begin{table}[ht!]
\addtolength{\tabcolsep}{-1.5pt}
\vspace*{-0.5em}
\centering
\small
\caption{Related Work on Foundational Time-Series Models.}
\vspace*{-0.5em}
\label{table:relatedwork}
\begin{tabular}{|c|c|c|}
\hline
\textit{Continual}  & \multicolumn{2}{c|}{\textit{Model Size}} \\ \cline{2-3}
\textit{Calibration} & \textit{Small} & \textit{Large} \\ \hline
 \cmark  & \texttt{TimeBlocks} & - \\ \hline
 \xmark  & \cite{abs-2401-03955,EkambaramJNSK23,ZhongSZLLC24,WangWSHLMZ024,0004Q00RP0G25,aurora_iclr,0004Q000RP0G25} & \cite{LiangZLWLP23,RuiHSKWW24,ZerveasJPBE21,YueWDYHTX22,ZuoLCCXB24,LiangCMIL24,pmlr-v235-woo24a,WuXWL21,NieNSK23} \\ \hline
\end{tabular}
\vspace{-1em}
\end{table}

Despite the existence of small models~\cite{abs-2401-03955,ZhongSZLLC24, WangWSHLMZ024}, they do not directly support zero-shot processing and require pre-definition of their input size. Also, models that provide zero-shot processing~\cite{GruverFQW23,ZhouNW0023,0011Q00C0HGZJ025} or manage multiple input sizes~\cite{pmlr-v235-woo24a,AshokMZCD24} still rely on very large models.
\texttt{TimeBlocks} addresses both issues by enabling the efficient construction of models with an automatic selection of input size.

\noindent
\textbf{Streaming Subset Building:}
The concept of compressing a data set into a representative subset, usually referred to as a coreset, has been extensively studied over the years~\cite{Phillips16,FeldmanL11,FeldmanT15,MaaloufEMFO24}. 
More recently, there has been a growing interest in building such coresets for scenarios involving data streams~\cite{BravermanFLRS23,MoSD24,TourHS24,LiSC22,BuzzegaBPAC20,CamposYKZGJ24} as it allows for the maintenance of a representation of the stream without requiring additional memory space.
These proposals have the drawback that they are specifically designed for particular problems, such as Support Vector Machines~\cite{TukanBFR21}, making them not suitable for other models, while others do not provide tight guarantees~\cite{Phillips16} or are not efficient for maintaining the coreset updated~\cite{LiSC22}. 
Thus, our proposal addresses all of these issues by introducing a low-cost coreset that only depends on the data stream, not the model.

\noindent
\textbf{Model Ensembling:}
The concept of combining models as a way to improve the performance of machine learning models has been studied for a long time. For example, ensembles~\cite{CaruanaNCK04,0002Z0KGJ23} and learned indexes~\cite{KraskaBCDP18} have been used for combining the results of several independent models. Approaches such as boosting~\cite{Schapire2003} and mixture-of-experts~\cite{AvnimelechI99} use the existing results of a model to establish the training of a new model. 
More recently, post hoc ensembles~\cite{PuruckerSABBH23} and routing experts~\cite{ZhouLLDHZDCLL22} select a combination of the best-performing models after training them on a particular task.
\texttt{TimeBlocks} differs from these methods by utilizing small pre-trained blocks that iteratively build a model. This approach eliminates the need for additional training while maintaining a small memory footprint. 
\section{Conclusion and Future Work} \label{conclusion}

The paper introduces \texttt{TimeBlocks}, a novel and efficient paradigm for building versatile, inference-assembled time-series models. By using small blocks pre-trained on various time-series tasks and data sets, \texttt{TimeBlocks} proposes a routing strategy to determine the most appropriate approach for processing a given time series during inference time. This includes selecting the optimal input size and choosing the most suitable blocks for processing a given time series. Additionally, \texttt{StreamCore}, a subset with approximation guarantees, is maintained under a data stream to periodically calibrate the model and ensure its reliability with minimal update cost. Experimental results demonstrate the effectiveness of the method for zero-shot time-series processing and calibration under a data stream, while handling different tasks.

In future work, it is of interest to extend the \texttt{Blockbase} to support a wider variety of blocks and compile statistics on their usage. This enhancement would enable support for more specialized models, evaluate quality of blocks, and improve scalability by optimizing both the size and usability of the \texttt{Blockbase}.

\bibliographystyle{ACM}
\bibliography{Sections/References}

\appendix
\setcounter{table}{0}
\renewcommand{\thetable}{A\arabic{table}}
\setcounter{figure}{0}
\renewcommand{\thefigure}{A\arabic{figure}}
\setcounter{algorithm}{0}
\renewcommand{\thealgorithm}{A\arabic{algorithm}}
\section{Appendix}

\subsection{Blocks Pre-Training} \label{appendix:block}
The process starts when receiving a pair of a time series with its corresponding label, along with the number of blocks required, as shown in line~\ref{line:pair}. In this pair, $x$ represents the time-series data in a window with a specific context length, while the label $y$ is the window of the next observed values over a specific horizon in forecasting. In imputation and outlier detection tasks, $y$ denotes the window of observed values with the same context length, while in classification $y$ is a label.
To build a model, $P$ blocks from a given architecture (e.g., MLP, LSTM, or Attention) are selected using their default configuration, and following a decreasing scale pattern (see Section~\ref{sssec:blockTraining}), as shown in lines~\ref{line:ModelStart}--\ref{line:ModelEnd}. Then, the model is trained minimizing a loss as shown in line~\ref{line:ModelTrain}.
Once the training is complete, the model and its blocks are returned.

\begin{algorithm}
\caption{Blocks Training.}
\label{alg:blocksTraining}
\begin{algorithmic}[1]

\Input : Time series pair $(x,y)$, number of blocks $P$ \label{line:pair}
\EndInput
\Output : Model, model blocks
\EndOutput
\vspace{0.5em}
\State $\mathit{Model} \gets \emptyset$
\For {$p $ in $ 1 \dots P$} \label{line:ModelStart}
    \State $B_{\mathit{new}} \gets$ New block with $2^{P-p}$ sampling.
    \State $\mathit{Model} \gets $ Attach $B_{\mathit{new}}$
\EndFor \label{line:ModelEnd}
\State Minimize $\mathit{Loss}(\mathit{Model}(x), y)$ \Comment{Model training} \label{line:ModelTrain}
\vspace{0.5em}
\State \Return $\mathit{Model}, \mathit{Model.Blocks}$ 
\end{algorithmic}
\end{algorithm}

\subsection{Router Training} \label{appendix:router}
Having a defined router and model, this process evaluates the model and records the residual of all blocks $r_p$, as shown in line~\ref{line:residuals}. Then, for each block of the model, it obtains the corresponding fingerprint (line~\ref{line:fingerprint}) and trains the router to approximate $Router(r_p) \approx b_{p+1}^1$, considering both the fingerprint and the output from the previous block, as shown in line~\ref{line:routerTrain}.
After iterating through all blocks, the router is returned to be trained with other end-to-end models. 

\begin{algorithm}
\caption{Router Training.}
\label{alg:routerTraining}
\begin{algorithmic}[1]

\Input : $\mathit{Router}(\cdot)$, $\mathit{Model}(\cdot)$
\EndInput
\Output : Updated $\mathit{Router}(\cdot)$
\EndOutput
\vspace{1em}
\State $\{r_1,\dots, r_P\} \gets $ Evaluate $\mathit{Model}$ \Comment{Record residuals} \label{line:residuals}
\For {$B_p$ in $\mathit{Model}$} \Comment{Iterate the blocks of the model}
    \State $b_{p}^1 \gets $ Get fingerprint from $B_p$ \label{line:fingerprint}
    \For{$e \gets 1,\dots,E$} \Comment{Epochs}
        \State Minimize $\mathit{Loss}(\mathit{Router}(r_{p-1}), b_{p}^1)$  \label{line:routerTrain} 
    \EndFor
\EndFor
\vspace{0.5em}
\State \Return $\mathit{Router}(\cdot)$

\end{algorithmic}
\end{algorithm}

\subsection{Block Selection} \label{appendix:router_eval}
Given a trained \texttt{Blockbase} and a new time series, the algorithm first evaluates the time series across different window sizes and then selects the most suitable cluster by comparing the their fingerprints, as shown in lines~\ref{line:contextStart}--\ref{line:contextEnds}.
Then, after selecting the processing cluster in line~\ref{line:blockSelection}, the time series is split accordingly and is standardized as explained in Section~\ref{ssec:processingblock}. Then, to add a new block $B_j$, the model is evaluated until the last selected block $B_{j-1}$, recording the associated residual $r_{j-1}$, as shown in line~\ref{line:registerHidden}. The residual is then evaluated by the router to compute the fingerprint for a candidate block in line~\ref{line:routerEval}.
The fingerprint is compared with the blocks within the cluster to find the most similar one by using vector similarity search~\cite{abs-2405-13954,JohnsonDJ21}, as shown in lines~\ref{line:ClusterStart}--\ref{line:ClusterEnd}.
Finally, the identified block is attached to the model. Having repeated the process to obtain up to $J$ blocks, the inference-time model is returned.

\begin{algorithm}
\caption{Block Selection.}
\label{alg:blockSelection}
\begin{algorithmic}[1]

\Input : Time series $x$, \texttt{Blockbase}, $\mathit{Router}(\cdot)$, block budget $J$
\EndInput
\Output : Zero-shot $\mathit{Model}$
\EndOutput
\vspace{0.5em}
\State $\mathit{Model} \gets \emptyset, \quad \mathit{Distances} \gets \emptyset$
\State $\mathit{Clusters} \gets $ Extract from \texttt{Blockbase}
\\
\For {$\mathit{Clus}$ in $\mathit{Clusters}$} \label{line:contextStart}
    \State $x_{st} \gets$ Split $x$ in windows of size $\mathit{Clus.ContextLength}$
    \State $r_{0} \gets $ Standardization $(x_{st})$ \Comment{Aligned with Sec.~\ref{sssec:blockTraining}}
    \State $\hat{B}_{cand}^1 \gets \mathit{Router}(r_{0})$  \Comment{Estimate candidate fingerprint}
    \State $\mathit{Distances}[\mathit{Clus}] \gets $ $\mathit{Distance}(\hat{B}_{\mathit{cand}}^1,\mathit{Clus.Fingerprint})$
\EndFor \label{line:contextEnds}
\vspace{0.5em}
\State $\mathit{minCluster} \gets \arg \min \mathit{Distances}$ \label{line:blockSelection}
\State $x_{st} \gets$ Split $x$ in windows of size of $\mathit{minCluster}$ center
\State $r_{0} \gets $ Standardization $(x_{st})$ \Comment{Aligned with Sec.~\ref{sssec:blockTraining}}
\\
\For {$j$ in $1,\dots,J$}
    \If {$j>1$}
    \State $r_{j-1} \gets $ Evaluate $\mathit{Model}(x_{st})$ \Comment{Register residual} \label{line:registerHidden}
    \EndIf
    \State $\hat{B}_{j}^1 \gets \mathit{Router}(r_{j-1})$  \Comment{Router estimates candidate fingerprint} \label{line:routerEval}
    \For {$\{B_{1},\dots, B_{P}\}$ in $\mathit{minCluster}$} \Comment{Blocks in the cluster} \label{line:ClusterStart}
        \State $\mathit{Distances}[B_{p}] \gets $ $\mathit{Distance}(\hat{B}_{j}^1,B_{p})$ \Comment{Similarity search}
    \EndFor \label{line:ClusterEnd}
    \State $\mathit{minBlock} \gets \arg \min \mathit{Distances}$
    \State $\mathit{Model} \gets $ Attach $\mathit{minBlock}$
\EndFor
\vspace{0.5em}
\State \Return $\mathit{Model}$

\end{algorithmic}
\end{algorithm}

\subsection{StreamCore Detailed Derivations} \label{appendix:streamcore}

Consider the following two properties:

\noindent
\textbf{Weight Relaxation:} Assume there is a mapping $\gamma: \mathcal{D} \rightarrow \mathcal{S}$ that assigns every data point $i \in \mathcal{D}$ to one $j \in \mathcal{S}$, therefore $\gamma(i)= j$. 

\noindent
\textbf{Lipschitz Continuous:} A function $f: X \rightarrow Y$ is $\lambda-$Lipschitz continuous if exists a constant $\lambda > 0$ such that for any $x_1,x_2 \in X$, $\lVert f(x_1)-f(x_2)\rVert \leq \lambda \lVert x_1-x_2 \rVert$. 

Then, considering weight relaxation, the sums of Equation~\ref{Eq2} in the main paper can be written as follows.
\begin{align}
    & \Bigg\lVert \sum^n_{i\in \mathcal{D}} l({x}_i,f(x_i)) - \sum^n_{j\in \mathcal{S}} w_j l({x}_j,f(x_j)) \Bigg\rVert \nonumber 
        \end{align}
    \begin{align}
    &= \left\lVert \sum^n_{i\in \mathcal{D}} l({x}_i,f(x_i)) - \sum^n_{i\in \mathcal{D}} l({x}_{\gamma(i)},f(x_{\gamma(i)})) \right\rVert \nonumber \\
    &\leq  \left\lVert \sum^n_{i\in \mathcal{D}} l({x}_i,f(x_i)) - l({x}_{\gamma(i)},f(x_{\gamma(i)})) \right\rVert, \label{Eq5}
\end{align}
where the last part of Equation~\ref{Eq5} is due to the triangle inequality and will be minimal when each point is assigned to the $j \in \mathcal{S}$ with the closest value. Therefore, we have
\begin{align}
    & \Bigg\lVert \sum^n_{i\in \mathcal{D}} l({x}_i,f(x_i)) - \sum^n_{j\in \mathcal{S}} w_j l({x}_j,f(x_j)) \Bigg\rVert \nonumber \\
    &\leq  \left\lVert \sum^n_{i\in \mathcal{D}} \min_{j \in \mathcal{S}} l({x}_i,f(x_i)) - l({x}_j,f(x_j)) \right\rVert \label{Eq6}
\end{align}

Assuming that both $f(\cdot)$ and $l(\cdot,\cdot)$ are Lipschitz Continuous, since $f$ is a deep learning based model and $l$ is a loss function, as derived by previous studies~\cite{LiSC22}, the right-hand side of Equation~\ref{Eq6} can be bounded as follows.
\begin{align}
    \lVert l({x}_i,f(x_i)) - l({x}_j,f(x_j)) \rVert & \leq \lambda_1 \lVert f(x_i) - f(x_j) \rVert \nonumber \\
    & \leq \lambda_1 \lambda_2 \lVert x_i - x_j \rVert = \lambda \lVert x_i - x_j\rVert \label{Eq8}
\end{align}

Applying the results of Equations~\ref{Eq6} and~\ref{Eq8} to Equation~\ref{Eq2}, we get:
\begin{gather}
    \frac{1}{n} \Bigg\lVert \sum^n_{i\in \mathcal{D}} l({x}_i,f(x_i)) - \sum^n_{j\in \mathcal{S}} w_j l({x}_j,f(x_j)) \Bigg\rVert \leq \frac{1}{n} \sum^n_{i\in \mathcal{D}} \min_{j \in \mathcal{S}} \lambda \lVert x_i - x_j \rVert \label{Eq9} \nonumber
\end{gather}
which is a bounded approximation for the coreset.

For computing the period $R$, it exists an $(8\alpha + 4)$-approximation with $R = \frac{\Phi}{4\alpha}$ updates~\cite{Cohen-AddadHPSS19}. 
Then, as each time $t$ the data stream includes $|\mathcal{D}^t|$ observations, the period is reduced to $R = \frac{\Phi}{4\alpha \times |\mathcal{D}^t|}$ as there are $|\mathcal{D}^t|$ updates every $t$. This means, for example, that a \texttt{StreamCore} of size $|\mathcal{S}| = 400$ can be updated after receiving data over 50 periods, as computed below.
\begin{gather}
    R = \frac{\Phi}{4\alpha \times |\mathcal{D}^t|} = \frac{|\mathcal{S}|\times|\mathcal{D}^t|}{4\alpha \times |\mathcal{D}^t|} = \frac{400}{4\times 2} = 50 \nonumber
\end{gather}

In the context of time-series data, a concept drift is treated as an external event that triggers model calibration. For example, when a change in the data stream exceeds a predefined threshold, the model can be calibrated using the \texttt{StreamCore}, a component that continuously collects new data, which may or may not include a concept drift. This approach enables a general evaluation, independent of when a concept drift occurs during deployment.

\subsection{Detailed Experimental Setup} \label{appendix:experimental_setup}
\subsubsection{Tasks and Data Sets}

We consider four different tasks: forecasting, imputation, outlier detection, and classification. For forecasting and imputations tasks we use popular forecasting data sets~\cite{ZhouZPZLXZ21}, namely \textit{ETTh1}, \textit{ETTh2}, \textit{ETTm1}, and \textit{ETTm2}, which measure Electricity Transformer Temperature at different frequencies, and \textit{Weather}, for meteorological data. For outlier detection, we use data sets related to web server monitoring (\textit{SMD}~\cite{SuZNLSP19}), spatial exploration (\textit{MSL}~\cite{HundmanCLCS18}), soil moisture (\textit{SMAP}~\cite{HundmanCLCS18}), and water treatment (\textit{SWaT}~\cite{MathurT16}). These data sets have ground truth labels that indicate outliers. We solely use these labels for evaluating model accuracy, not for model training.
Table~\ref{table:data_sets} presents the details for the data sets, including their number of dimensions, training partition, and frequency. For classification, we use the UCR time-series archive~\cite{DauBKYZGRK19}, which consists of 128 data sets from a wide variety of domains.

For all tasks, the context length for the baselines is set to 96, as it is used most commonly~\cite{WuHLZ0L23, WuXWL21, ZhouMWW0022}, while for pre-training the blocks, we consider a power-of-two set of context lengths varying $2^6$ to $2^{11}$, i.e., $\{64, 128, 256, 512, 1024, 2048\}$.

\subsubsection{Metrics}

To evaluate performance on forecasting and imputations tasks, we consider two metrics: Mean Absolute Error (\textit{MAE}) and Mean Squared Error (\textit{MSE}) computed over the testing set after implementing either the inference-time or calibrated model. These metrics are defined as follows.
\begin{gather}
    \mathit{MAE} = \frac{1}{n}\sum_{i=1}^n | y - \hat{y} | \qquad \mathit{MSE} = \frac{1}{n}\sum_{i=1}^n ( y - \hat{y} )^2 \nonumber
\end{gather}

For the outlier detection task, we use the F1 Score metric, which is the harmonic mean of precision and recall, with a self-determined threshold~\cite{GoswamiCCMK23}.

For classification, we generate a critical difference diagram based on \textit{Accuracy} after applying the null-hypothesis Friedman test~\cite{Friedman1940ACO} and the Wilcoxon–Holm post-hoc test~\cite{Holm,Wilcoxon} to rank the evaluated methods. The diagram is built by computing the difference in performance between the baselines that support classification for each data set and ranking them from the smallest to the largest difference. Then, the average rank across all data sets is calculated, with a thick horizontal line indicating when the difference between the evaluated methods is not statistically significant.

\subsubsection{Baselines}

To evaluate the effectiveness of the the proposal, we compare it with the following baselines from three different categories to enable a fair comparison among different time-series processing approaches:

$\text{  }$

\noindent
\textbf{Time-Series Specialized Models:}
\begin{itemize}
    \item \texttt{Autoformer}~\cite{WuXWL21}: a Transformer-based model that introduces time-series decomposition through an auto-correlation mechanism to handling time-dependency, instead of exclusively relying on default language processing.
    \item \texttt{TimesNet}~\cite{WuHLZ0L23}: introduces a two-dimensional decomposition of time series to measure variations both between periods and within a given period. This approach allows for the use of two-dimensional methods to effectively process time-series data.
    \item \texttt{PatchTST}~\cite{NieNSK23}: a method that segments time series into sub-series called patches. This allows for processing them as contextualized tokens rather than as independent observations within a Transformer-based encoding mechanism.
    \item \texttt{AnomalyTransformer}~\cite{XuWWL22}: a popular Transformer-based outlier detection method enhanced by a metric measuring the associativity between adjacent data points.
    \item \texttt{DCdetector}~\cite{YangZZW023}: a representation learning outlier detection method uses multi-scale attention layers to identify weak correlations among data points.
    \item \texttt{InceptionTime}~\cite{FawazLFPSWWIMP20}: a convolutional neural network that employs variable-length convolution filters to capture temporal patterns across different timescales.
    \item \texttt{ResNet}~\cite{WangYO17}: an adaptation of the backbone residual network architecture for time-series classification.
\end{itemize}
\textbf{LLM-Based Model:}
\begin{itemize}
    \item \texttt{Lag-Llama}~\cite{Rasul23}: introduces the architecture of an LLM~\cite{abs-2302-13971} for processing time series while considering the covariance of each observation with respect to its context length.
\end{itemize}
\textbf{Time-Series Foundational Models:}
\begin{itemize}
    \item \texttt{TimeMixer}~\cite{WangWSHLMZ024}: proposes granular time-series decomposition and then mixing the results in order to identify time-series properties such as trends and cycles. This approach, called a mixer, enables more contextualized processing of time series.
    \item \texttt{Tiny Time Mixers (TTM)}~\cite{abs-2401-03955}: unifies existing studies~\cite{ZhongSZLLC24,EkambaramJNSK23}, bringing an adaptive patching and a mixer architecture into an efficient MLP structure to enable efficient time-series processing.
    \item \texttt{Moment}~\cite{GoswamiSCCLD24}: combines the patching mechanism with a Transformer encoder, which are then trained using a masking strategy to facilitate the processing of multiple tasks.
    \item \texttt{SymTime}~\cite{wang2025synthetic}: it introduces the pairing of time series with symbolic expressions to overcome data scarcity. It is only evaluated for reference in the forecasting task (i.e. Table~\ref{table:baselines_zero}) due to its parallel release.
\end{itemize}

Potential baselines such \texttt{Moirai}~\cite{WooLKXSS24}, \texttt{ToTo}~\cite{abs-2407-07874}, and \texttt{Chronos}~\cite{AnsariSTZMSSRPK24} are not evaluated, as they are specifically designed for time\Hyphdash series forecasting, requiring significant adaptation and retraining to function in other tasks.

\subsubsection{Implementation}
The proposed method is implemented using Python 3.9.19 and the deep learning framework PyTorch 2.3.1. All models are tested under Ubuntu 22.04.4 using Titan RTX GPUs with 24GB VRAM and an Intel Xeon W-2155 with 128GB RAM.
The code is available as supplementary material. 
When applicable, all methods use the hyper-parameter configurations recommended in their documentation. Otherwise, we adjust them using the validation set, following common practices. For \texttt{Lag-Llama} we use the average of 100 samples from its distribution. When training the blocks and the router, we use the AdamW optimizer~\cite{LoshchilovH19} trained for 50 epochs with a learning rate of $0.001$ and a batch size of 64.

To build the inference-time model, we utilize the router to select $J=6$ blocks (see Algorithm~\ref{alg:blockSelection}) as the default backbone, and we also vary it, as will be discussed in Section~\ref{sssec_apendix:effectBlocks}. This selection is to ensure consistency and comparability in terms of efficiency when compared to the efficient \texttt{TTM} baseline. 
The comparison between the block candidate and the \texttt{TimeBlocks} fingerprints is calculated using cosine distance. For all the evaluations, the best results are highlighted in bold, while the second best results are underlined.

\begin{table}[ht!]
\addtolength{\tabcolsep}{-1pt}
\vspace{-1em}
    \small
    \centering
    \caption{Data Sets.}
    \label{table:data_sets}
    \begin{tabular}{ |l|l|c|c|c|c| } 
    \hline
    &\textbf{Data Set} &  \textit{Dims} & \textit{Time Steps} & \textit{Train/Valid/Test} & \textit{Frequency} \\
    \hline
    \parbox[t]{2mm}{\multirow{5}{*}{\rotatebox[origin=c]{90}{\textit{Forecasting}}}} & \textit{ETTh1} & 7 & 17420 & 10452/3484/3484 & Hours \\
    & \textit{ETTh2} & 7 & 17420 & 10452/3484/3484 & Hours \\
    & \textit{ETTm1} & 7 & 69680 & 41808/13936/13936 & Minutes \\
    & \textit{ETTm2} & 7 & 69680 & 41808/13936/13936 & Minutes \\
    & \textit{Weather} & 21 & 52696 & 24388/14154/14154 & Hours \\
    \hline
    \parbox[t]{2mm}{\multirow{4}{*}{\rotatebox[origin=c]{90}{\textit{Outlier}}}} & \textit{SMD} & 25 & 7.18 million & 4.31/1.43/1.43 million & Seconds \\
    & \textit{MSL} & 55 & 132046 & 58318/36864/36864 & Hours \\
    & \textit{SMAP} & 38 & 562800 & 135184/213808/213808 & Hours \\
    & \textit{SWaT} & 51 & 489600 & 293760/97920/97920 & Seconds \\
    \hline
    \end{tabular}
    \vspace{-1em}
\end{table}

\subsection{Additional Experimental Results} \label{appendix:experiments}

\subsubsection{Disaggregated Results} 
Detailed results for Tables \ref{table:baselines_zero}, \ref{table:baselines_continual}, and \ref{table:baselines_imputation} in the main paper are shown in Tables \ref{table:baselines_zeroshot}, \ref{table:baselines_finetuning}, and \ref{table_appendix:baselines_imputation}, respectively.

\begin{table}[ht!]
    \small
    \vspace{-1em}
    \centering
    \caption{Imputation Inference-Time Model (\textit{MSE}). Detailed results for Table~\ref{table:baselines_imputation}.}
    \label{table_appendix:baselines_imputation}
    \begin{tabular}{ |l|*{6}{c|} } 
    \hline

\textbf{Model} & \textit{Mask} & \textit{ETTh1} & \textit{ETTh2} & \textit{ETTm1} & \textit{ETTm2} & \textit{Weather} \\
 \hline \multirow{3}*{\texttt{Autoformer}}  
 & 10\% & 0.549 & 0.365 & 0.467 & 0.280 & 0.326 \\ 
 & 20\% & 0.807 & 0.488 & 0.591 & 0.417 & 0.346 \\ 
 & 30\% & 1.313 & 0.552 & 0.834 & 0.488 & 0.423 \\ 
 \hline \multirow{3}*{\texttt{PatchTST}}  
 & 10\% & 0.427 & 0.307 & 0.323 & 0.246 & 0.203 \\ 
 & 20\% & 0.618 & 0.317 & 0.398 & 0.260 & 0.281 \\ 
 & 30\% & 0.655 & 0.380 & 0.542 & 0.416 & 0.358 \\ 
 \hline \multirow{3}*{\texttt{TimeMixer}}  
 & 10\% & 0.438 & 0.294 & 0.319 & 0.240 & \underline{0.201} \\ 
 & 20\% & 0.560 & 0.375 & \textbf{0.395} & 0.262 & 0.265 \\ 
 & 30\% & 1.148 & 0.375 & 0.605 & 0.401 & 0.342 \\ 
 \hline \multirow{3}*{\texttt{TimesNet}}  
 & 10\% & 0.441 & 0.324 & 0.340 & 0.253 & 0.223 \\ 
 & 20\% & 0.723 & 0.326 & 0.415 & 0.308 & 0.279 \\ 
 & 30\% & 0.787 & 0.411 & 0.704 & 0.417 & 0.356 \\ 
 \hline \multirow{3}*{\texttt{Lag-Llama}}  
 & 10\% & 0.427 & 0.290 & 0.311 & 0.234 & 0.205 \\ 
 & 20\% & 0.616 & 0.324 & 0.405 & 0.287 & 0.272 \\ 
 & 30\% & 0.839 & 0.399 & 0.604 & 0.380 & 0.322 \\ 
  \hline \multirow{3}*{\texttt{Moment}} 
 & 10\% & 0.497 & 0.331 & 0.404 & 0.279 & 0.254 \\ 
 & 20\% & 0.741 & 0.418 & 0.501 & 0.354 & 0.325 \\ 
 & 30\% & 1.038 & 0.463 & 0.749 & 0.456 & 0.411 \\ 
 \hline \multirow{3}*{\texttt{TTM}}  
 & 10\% & \underline{0.417} & \underline{0.234} & \underline{0.260} & \underline{0.186} & 0.202 \\ 
 & 20\% & \underline{0.539} & \underline{0.307} & 0.406 & \underline{0.258} & \underline{0.245} \\ 
 & 30\% & \underline{0.627} & \underline{0.361} & \underline{0.514} & \underline{0.307} & \underline{0.287} \\ 
 \hline
 \multirow{3}*{\texttt{TimeBlocks}}  
 & 10\% & \textbf{0.406} & \textbf{0.231} & \textbf{0.251} & \textbf{0.184} & \textbf{0.193} \\ 
 & 20\% & \textbf{0.529} & \textbf{0.304} & \underline{0.397} & \textbf{0.256} & \textbf{0.240} \\ 
 & 30\% & \textbf{0.617} & \textbf{0.357} & \textbf{0.506} & \textbf{0.304} & \textbf{0.270} \\ 
 \hline
\end{tabular}
\vspace{-0.5em}
\end{table}

\begin{table*}[ht!]
    \small
    \centering
    \caption{Forecasting Inference-Time Model. Detailed results for Table~\ref{table:baselines_zero}.}
    \label{table:baselines_zeroshot}
    \begin{tabular}{ |l|*{17}{c|} } 
    \hline

\multicolumn{2}{|c}{\textbf{Model}} & \multicolumn{2}{|c}{\texttt{Autoformer}} & \multicolumn{2}{|c}{\texttt{PatchTST}} & \multicolumn{2}{|c}{\texttt{TimeMixer}} & \multicolumn{2}{|c}{\texttt{TimesNet}} & \multicolumn{2}{|c}{\texttt{Lag-Llama}} & \multicolumn{2}{|c}{\texttt{Moment}} & \multicolumn{2}{|c}{\texttt{TTM}} & \multicolumn{2}{|c|}{\texttt{TimeBlocks}} \\
 \hline
\textit{Data Set} & \textit{Horizon} & \textit{MAE} & \textit{MSE} & \textit{MAE} & \textit{MSE} & \textit{MAE} & \textit{MSE} & \textit{MAE} & \textit{MSE} & \textit{MAE} & \textit{MSE} & \textit{MAE} & \textit{MSE} & \textit{MAE} & \textit{MSE} & \textit{MAE} & \textit{MSE} \\
 \hline \multirow{4}*{\textit{ETTh1}}  
 & 96 & 0.714 & 1.018 & 0.521 & 0.586 & 0.467 & 0.500 & 0.596 & 0.742 & 0.557 & 0.647 & 0.455 & 0.471 & \underline{0.404} & \textbf{0.365} & \textbf{0.397} & \underline{0.370} \\ 
 & 192 & 0.638 & 0.847 & 0.578 & 0.719 & 0.505 & 0.560 & 0.600 & 0.744 & 0.563 & 0.703 & 0.474 & 0.505 & \underline{0.429} & \textbf{0.393} & \textbf{0.425} & \underline{0.409} \\ 
 & 336 & 0.715 & 1.022 & 0.583 & 0.715 & 0.543 & 0.655 & 0.646 & 0.846 & 0.597 & 0.745 & 0.491 & 0.534 & \underline{0.448} & \textbf{0.415} & \textbf{0.442} & \underline{0.429} \\ 
 & 720 & 0.770 & 1.149 & 0.582 & 0.681 & 0.751 & 1.148 & 0.640 & 0.837 & 0.638 & 0.906 & 0.515 & \underline{0.538} & \underline{0.493} & 0.538 & \textbf{0.475} & \textbf{0.456} \\ 
 \hline \multirow{4}*{\textit{ETTh2}} 
 & 96 & 0.535 & 0.532 & 0.349 & 0.302 & 0.345 & 0.296 & 0.375 & 0.334 & 0.397 & 0.352 & 0.519 & 0.612 & \underline{0.343} & \underline{0.285} & \textbf{0.337} & \textbf{0.271} \\ 
 & 192 & 0.553 & 0.575 & 0.395 & 0.378 & 0.389 & 0.375 & 0.424 & 0.424 & 0.409 & 0.416 & 0.535 & 0.649 & \underline{0.380} & \underline{0.341} & \textbf{0.378} & \textbf{0.333} \\ 
 & 336 & 0.579 & 0.624 & 0.432 & 0.424 & 0.420 & 0.410 & 0.447 & 0.448 & 0.455 & 0.458 & 0.560 & 0.670 & \underline{0.404} & \underline{0.383} & \textbf{0.401} & \textbf{0.355} \\ 
 & 720 & 0.577 & 0.616 & 0.450 & 0.435 & \textbf{0.431} & \underline{0.409} & 0.461 & 0.456 & 0.441 & 0.450 & 0.593 & 0.689 & 0.439 & 0.441 & \underline{0.434} & \textbf{0.392} \\ 
 \hline \multirow{4}*{\textit{ETTm1}}  
 & 96 & 0.618 & 0.808 & 0.451 & 0.516 & 0.476 & 0.570 & 0.540 & 0.709 & 0.485 & 0.612 & 0.564 & 0.710 & \underline{0.357} & \underline{0.413} & \textbf{0.343} & \textbf{0.288} \\ 
 & 192 & 0.600 & 0.772 & 0.477 & 0.542 & 0.478 & 0.550 & 0.566 & 0.736 & 0.525 & 0.637 & 0.576 & 0.726 & \underline{0.374} & \underline{0.476} & \textbf{0.368} & \textbf{0.326} \\ 
 & 336 & 0.644 & 0.872 & 0.543 & 0.691 & 0.501 & 0.605 & 0.553 & 0.681 & 0.515 & 0.698 & 0.592 & 0.738 & \underline{0.398} & \underline{0.553} & \textbf{0.386} & \textbf{0.362} \\ 
 & 720 & 0.637 & 0.840 & 0.591 & 0.767 & 0.525 & \underline{0.615} & 0.599 & 0.826 & 0.582 & 0.693 & 0.612 & 0.742 & \underline{0.417} & 0.737 & \textbf{0.413} & \textbf{0.401} \\ 
 \hline \multirow{4}*{\textit{ETTm2}}  
 & 96 & 0.480 & 0.420 & 0.272 & 0.193 & 0.277 & 0.201 & 0.304 & 0.226 & 0.330 & 0.244 & 0.553 & 0.695 & \underline{0.257} & \underline{0.187} & \textbf{0.257} & \textbf{0.164} \\ 
 & 192 & 0.482 & 0.457 & 0.311 & \underline{0.254} & 0.315 & 0.262 & 0.356 & 0.310 & 0.351 & 0.318 & 0.566 & 0.717 & \underline{0.302} & 0.261 & \textbf{0.295} & \textbf{0.216} \\ 
 & 336 & 0.519 & 0.524 & 0.345 & \underline{0.308} & 0.351 & 0.320 & 0.379 & 0.349 & 0.390 & 0.357 & 0.574 & 0.723 & \underline{0.331} & 0.323 & \textbf{0.328} & \textbf{0.259} \\ 
 & 720 & 0.560 & 0.605 & 0.412 & \underline{0.416} & 0.411 & 0.423 & 0.420 & 0.434 & 0.433 & 0.460 & 0.584 & 0.703 & \underline{0.384} & 0.436 & \textbf{0.378} & \textbf{0.329} \\ 
 \hline \multirow{4}*{\textit{Weather}} 
 & 96 & 0.340 & 0.272 & 0.212 & 0.168 & 0.207 & 0.159 & 0.219 & 0.168 & 0.225 & 0.184 & 0.560 & 0.717 & \underline{0.202} & \underline{0.154} & \textbf{0.194} & \textbf{0.145} \\ 
 & 192 & 0.401 & 0.352 & 0.251 & 0.214 & 0.249 & 0.204 & 0.265 & 0.224 & 0.271 & 0.229 & 0.572 & 0.734 & \underline{0.245} & \underline{0.203} & \textbf{0.238} & \textbf{0.191} \\ 
 & 336 & 0.399 & 0.360 & 0.295 & 0.272 & \underline{0.287} & 0.257 & 0.305 & 0.283 & 0.318 & 0.278 & 0.590 & 0.743 & 0.287 & \underline{0.256} & \textbf{0.278} & \textbf{0.243} \\ 
 & 720 & 0.431 & 0.429 & 0.345 & 0.347 & 0.342 & 0.335 & 0.354 & 0.355 & \underline{0.339} & 0.348 & 0.603 & 0.731 & 0.348 & \underline{0.329} & \textbf{0.330} & \textbf{0.302} \\ 
 \hline
\end{tabular}
\end{table*}

\begin{table*}[ht!]
\addtolength{\tabcolsep}{-1.2pt}
    \small
    \centering
    \caption{Forecasting Continual Model. Subset Size 5\%. Detailed results for Table~\ref{table:baselines_continual}.}
    \label{table:baselines_finetuning}
    \begin{tabular}{ |l|*{19}{c|} } 
    \hline

\multicolumn{2}{|c}{\multirow{2}{*}{\textbf{Model}}} & \multicolumn{2}{|c}{\multirow{2}{*}{\texttt{Autoformer}}} & \multicolumn{2}{|c}{\multirow{2}{*}{\texttt{PatchTST}}} & \multicolumn{2}{|c}{\multirow{2}{*}{\texttt{TimeMixer}}} & \multicolumn{2}{|c}{\multirow{2}{*}{\texttt{TimesNet}}} & \multicolumn{2}{|c}{\multirow{2}{*}{\texttt{Lag-Llama}}} & \multicolumn{2}{|c}{\multirow{2}{*}{\texttt{Moment}}} & \multicolumn{2}{|c}{\multirow{2}{*}{\texttt{TTM}}} & \multicolumn{2}{|c}{\multirow{2}{*}{\texttt{TimeBlocks}}} & \multicolumn{2}{|c|}{\texttt{TimeBlocks}} \\
\multicolumn{2}{|c}{} & \multicolumn{2}{|c}{} & \multicolumn{2}{|c}{} & \multicolumn{2}{|c}{} & \multicolumn{2}{|c}{} & \multicolumn{2}{|c}{} & \multicolumn{2}{|c}{} & \multicolumn{2}{|c}{} & \multicolumn{2}{|c}{} & \multicolumn{2}{|c|}{\texttt{+ Random}} \\
 \hline
\textit{Data Set} & \textit{Horizon} & \textit{MAE} & \textit{MSE} & \textit{MAE} & \textit{MSE} & \textit{MAE} & \textit{MSE} & \textit{MAE} & \textit{MSE} & \textit{MAE} & \textit{MSE} & \textit{MAE} & \textit{MSE} & \textit{MAE} & \textit{MSE} & \textit{MAE} & \textit{MSE} & \textit{MAE} & \textit{MSE} \\
 \hline \multirow{4}*{\textit{ETTh1}}  
 & 96 & 0.501 & 0.549 & 0.402 & 0.381 & 0.397 & 0.375 & 0.420 & 0.399 & \textbf{0.396} & 0.396 & 0.422 & 0.414 & 0.400 & 0.366 & \underline{0.397} & \textbf{0.356} & 0.401 & \underline{0.362} \\ 
 & 192 & 0.499 & 0.525 & 0.439 & 0.432 & 0.429 & 0.438 & 0.448 & 0.447 & \textbf{0.422} & 0.424 & 0.450 & 0.463 & 0.423 & \underline{0.391} & \underline{0.423} & \textbf{0.389} & 0.448 & 0.397 \\ 
 & 336 & 0.502 & 0.522 & 0.472 & 0.490 & 0.462 & 0.507 & 0.462 & 0.477 & 0.470 & 0.494 & 0.464 & 0.493 & \underline{0.427} & 0.421 & \textbf{0.418} & \textbf{0.377} & 0.455 & \underline{0.390} \\ 
 & 720 & 0.516 & 0.519 & 0.522 & 0.555 & 0.487 & 0.523 & 0.491 & 0.506 & 0.499 & 0.520 & \underline{0.477} & 0.476 & 0.490 & 0.538 & \textbf{0.474} & \textbf{0.450} & 0.479 & \underline{0.467} \\ 
 \hline \multirow{4}*{\textit{ETTh2}} 
 & 96 & 0.452 & 0.426 & 0.359 & 0.312 & 0.345 & 0.294 & 0.372 & 0.325 & 0.351 & 0.326 & 0.497 & 0.558 & \underline{0.340} & 0.282 & \textbf{0.334} & \textbf{0.271} & 0.348 & \underline{0.273} \\ 
 & 192 & 0.488 & 0.480 & 0.407 & 0.385 & 0.393 & 0.374 & 0.418 & 0.401 & 0.401 & 0.394 & 0.511 & 0.597 & \underline{0.379} & 0.338 & \textbf{0.377} & \textbf{0.330} & 0.396 & \underline{0.332} \\ 
 & 336 & 0.518 & 0.529 & 0.429 & 0.410 & 0.445 & 0.438 & 0.447 & 0.435 & 0.441 & 0.421 & 0.537 & 0.629 & \underline{0.403} & 0.383 & \textbf{0.401} & \textbf{0.354} & 0.417 & \underline{0.356} \\ 
 & 720 & 0.566 & 0.645 & 0.456 & 0.435 & 0.451 & 0.440 & 0.465 & 0.457 & 0.465 & 0.450 & 0.559 & 0.622 & \underline{0.435} & 0.441 & \textbf{0.433} & \textbf{0.381} & 0.459 & \underline{0.387} \\ 
 \hline \multirow{4}*{\textit{ETTm1}}  
 & 96 & 0.484 & 0.512 & 0.355 & 0.310 & 0.359 & 0.319 & 0.372 & 0.334 & 0.365 & 0.365 & 0.553 & 0.703 & \underline{0.331} & 0.359 & \textbf{0.323} & \textbf{0.289} & 0.333 & \underline{0.295} \\ 
 & 192 & 0.526 & 0.603 & 0.385 & 0.360 & 0.383 & 0.360 & 0.413 & 0.413 & 0.418 & 0.430 & 0.565 & 0.717 & \underline{0.371} & 0.402 & \textbf{0.366} & \textbf{0.329} & 0.398 & \underline{0.332} \\ 
 & 336 & 0.524 & 0.601 & 0.399 & 0.381 & 0.409 & 0.395 & 0.424 & 0.418 & 0.421 & 0.422 & 0.581 & 0.726 & 0.384 & 0.424 & \textbf{0.368} & \textbf{0.343} & \underline{0.383} & \underline{0.352} \\ 
 & 720 & 0.545 & 0.632 & 0.439 & 0.444 & 0.439 & 0.450 & 0.447 & 0.460 & 0.440 & 0.462 & 0.603 & 0.725 & \underline{0.416} & 0.575 & \textbf{0.411} & \textbf{0.406} & 0.440 & \underline{0.411} \\ 
 \hline \multirow{4}*{\textit{ETTm2}}  
 & 96 & 0.326 & 0.243 & 0.265 & 0.178 & 0.259 & 0.174 & 0.268 & 0.187 & 0.277 & 0.194 & 0.546 & 0.685 & \underline{0.248} & 0.174 & \textbf{0.246} & \textbf{0.161} & 0.264 & \underline{0.163} \\ 
 & 192 & 0.347 & 0.289 & 0.307 & 0.243 & 0.301 & 0.240 & 0.307 & 0.246 & 0.293 & 0.239 & 0.557 & 0.708 & \underline{0.288} & 0.240 & \textbf{0.287} & \textbf{0.215} & 0.299 & \underline{0.219} \\ 
 & 336 & 0.394 & 0.373 & 0.345 & 0.301 & 0.340 & 0.297 & 0.348 & 0.313 & \underline{0.328} & 0.305 & 0.567 & 0.715 & 0.328 & 0.299 & \textbf{0.326} & \textbf{0.264} & 0.339 & \underline{0.267} \\ 
 & 720 & 0.426 & 0.432 & 0.401 & 0.397 & 0.398 & 0.401 & 0.403 & 0.410 & 0.395 & 0.394 & 0.578 & 0.694 & \underline{0.381} & 0.407 & \textbf{0.375} & \textbf{0.336} & 0.397 & \underline{0.346} \\ 
 \hline \multirow{4}*{\textit{Weather}} 
 & 96 & 0.331 & 0.249 & 0.206 & 0.160 & 0.205 & 0.155 & 0.219 & 0.168 & 0.228 & 0.168 & 0.554 & 0.710 & \underline{0.196} & 0.152 & \textbf{0.192} & \textbf{0.144} & 0.198 & \underline{0.145} \\ 
 & 192 & 0.381 & 0.326 & 0.243 & 0.203 & 0.248 & 0.201 & 0.264 & 0.223 & 0.272 & 0.229 & 0.569 & 0.729 & 0.240 & 0.198 & \textbf{0.231} & \textbf{0.190} & \underline{0.235} & \underline{0.191} \\ 
 & 336 & 0.384 & 0.346 & 0.290 & 0.262 & 0.280 & 0.249 & 0.302 & 0.279 & 0.292 & 0.264 & 0.583 & 0.732 & \underline{0.275} & 0.250 & \textbf{0.269} & \textbf{0.242} & 0.293 & \underline{0.243} \\ 
 & 720 & 0.422 & 0.423 & 0.340 & 0.336 & 0.340 & 0.327 & 0.353 & 0.353 & \underline{0.331} & 0.331 & 0.594 & 0.719 & 0.338 & 0.326 & \textbf{0.325} & \textbf{0.305} & 0.338 & \underline{0.321} \\ 
 \hline
\end{tabular}
\end{table*}

\subsubsection{Effect of the Number Blocks} \label{sssec_apendix:effectBlocks}

Even when the linear models are significantly smaller~\cite{ZengCZ023}, which allows for the construction of very efficient models, as enabled by \texttt{TimeBlocks}, we want to assess the efficiency of the router in selecting a specific number of blocks $J$ for a model, and how that affects the performance. Therefore, when building a model, we evaluate the impact of adding a new block and evaluating the model performance up to that point.

\begin{figure}[ht]
\small
\centering
\begin{subfigure}{0.5\linewidth}
    \begin{tikzpicture}
        \begin{axis}[
    xlabel=Number of blocks ($J$),
    ylabel=Normalized MSE,
            width=1*\linewidth,
            height=0.5*\axisdefaultheight,
            legend image post style={xscale=0.4},
            legend style={
                    at={(-0.2,1.57)},
                    anchor=north west,
                    legend columns=3,},font=\footnotesize]
            \addplot[blue,mark=none] table[x=Block, y=ETTh1] {Figures/Data/BlocksNorm.txt};
            \addlegendentry{\textit{ETTh1}}
            \addplot[red,mark=none] table[x=Block, y=ETTh2] {Figures/Data/BlocksNorm.txt};
            \addlegendentry{\textit{ETTh2}}
            \addplot[black,mark=none] table[x=Block, y=ETTm1] {Figures/Data/BlocksNorm.txt};
            \addlegendentry{\textit{ETTm1}}
            \addplot[purple,dashed, mark=none] table[x=Block, y=ETTm2] {Figures/Data/BlocksNorm.txt};
            \addlegendentry{\textit{ETTm2}}
            \addplot[blue,dashed,mark=none] table[x=Block, y=Weather] {Figures/Data/BlocksNorm.txt};
            \addlegendentry{\textit{Weather}}
        \end{axis}
    \end{tikzpicture}
    \caption{Normalized Error.}
    \label{subfig:blocksNorm}
\end{subfigure}
\hspace*{-1.5ex}
\begin{subfigure}{0.5\linewidth}
    \begin{tikzpicture}
        \begin{axis}[
            xlabel=Number of blocks ($J$),
            ylabel=MSE,
            width=1*\linewidth,
            height=0.5*\axisdefaultheight,
            legend style={
                    at={(-0.1,1.4)},
                    anchor=north west,
                    legend columns=2,},font=\footnotesize]
            \addplot[red,mark=*] table[x=Block, y=ETTh2] {Figures/Data/BlocksAbs.txt};
            \addlegendentry{\textit{ETTh2}}
            \addplot[black,mark=*] table[x=Block, y=ETTm1] {Figures/Data/BlocksAbs.txt};
            \addlegendentry{\textit{ETTm1}}
        \end{axis}
    \end{tikzpicture}
    \caption{Non-normalized Error.}
    \label{subfig:blocksAbs}
\end{subfigure}
\caption{Model Size Efficiency. Forecasting horizon is 96.}
\label{fig:blocks}
\Description[BlockSelection]{BlockSelection}
\end{figure}
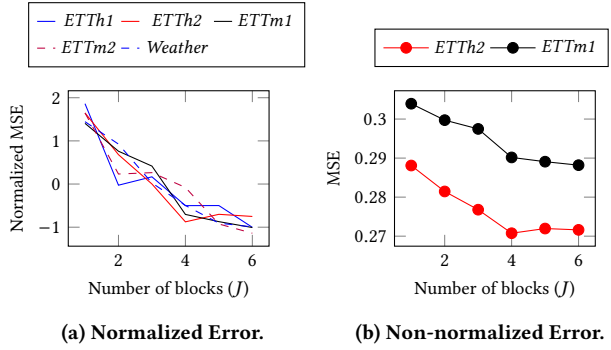

The results in Figure~\ref{fig:blocks} show that adding new blocks improves performance. In Figure~\ref{subfig:blocksNorm}, the results are normalized to show the trend among data sets, indicating that when adding new blocks, the normalized error decreases. Further analysis in Figure~\ref{subfig:blocksAbs} for two data sets reveals that the absolute change is relatively low when new blocks are added. This suggests that a model with few blocks, such as $J=4$, can achieve similar performance to a larger model, improving the efficiency of the model proposal.

\subsubsection{Pre-Training Cost} \label{sssec_apendix:pre-training}
The pretraining of a large number of blocks in the \texttt{BlockBase} is, in principle, comparable to training a single large foundational model, with two significant differences. First, the blocks can be pretrained independently and in parallel, which speeds up training compared to training a single large model. For example, a training step for a six-block model takes less than second, while the same step in the \texttt{Lag-Llama} model will take more than a minute, Second, during inference, as shown in Figure~\ref{subfig:runtime}, the \texttt{TimeBlocks} approach enables the deployment of lightweight and efficient models, while the large-model approach requires loading the entire large model, leading to considerably higher computational costs. 

\subsubsection{Coreset Evaluation} 
To extend the results shown in the last two rows of Table~\ref{table:baselines_continual}, we evaluate 
\texttt{StreamCore} with two coreset methods: \texttt{DER++}~\cite{BuzzegaBPAC20}, an experience replay strategy, and \texttt{Camel}~\cite{LiSC22}, which implements a buffer. The results for three forecasting data sets in Table~\ref{table:coresets} show that \texttt{StreamCore} maintains a high performance, mainly due to its tight guarantee.

\begin{table}[ht!]
\addtolength{\tabcolsep}{-1pt}
    \small
    \centering
    \caption{Forecasting Continual Model With Coresets (Average MSE).}
    \label{table:coresets}
    \begin{tabular}{ |l|*{3}{c|} } 
    \hline
    \textbf{Coreset}
    & \textit{ETTh1} & \textit{ETTh2} & \textit{Weather} \\
    \hline
\texttt{DER++} & 0.419 & 0.517 & 0.346 \\ 
\texttt{Camel} & 0.450 & 0.443 & 0.312 \\ 
\texttt{StreamCore}& \textbf{0.393} & \textbf{0.342} & \textbf{0.220} \\ 

\hline
\end{tabular}
\end{table}

\subsubsection{Statistical Breakdown of Block Usage} 
To evaluate the \texttt{Block-} \texttt{base} usage, a statistical breakdown is shown in Table~\ref{table:statistical_break}. It considers a sample of five hundred blocks across the four evaluated tasks and related data sets, disaggregated by context length. Thus, each row represents the percentage usage of each particular block type for a given context length.

\begin{table}[hb!]
    \small
    \centering
    \caption{Statistical Breakdown of Block Usage by Task and Context Length (Block Use Percentage).}
    \label{table:statistical_break}
    \begin{tabular}{ |l|*{3}{c|} }
    \hline
  \textbf{Task (Data set)} & \multicolumn{3}{c|}{\textit{Block Type}} \\
  \cline{2-4}
  \textbf{-- Context Length} & \textit{Attention} & \textit{LSTM} & \textit{MLP} \\
  \hline 
\multicolumn{4}{|l|}{\textit{Classification (Adiac)}}  \\
\hline
-- 256 & 0.313 & 0.000 & 0.688 \\
-- 512 & 0.667 & 0.278 & 0.056 \\
-- 1024 & 0.000 & 0.000 & 1.000 \\
-- 2048 & 0.375 & 0.063 & 0.563 \\
\hline
\multicolumn{4}{|l|}{\textit{Forecasting (ETTh1, ETTh2)}}  \\
\hline
-- 256 & 0.269 & 0.321 & 0.410 \\
-- 512 & 0.141 & 0.218 & 0.641 \\
-- 1024 & 0.371 & 0.258 & 0.371 \\
-- 2048 & 0.238 & 0.286 & 0.476 \\
\hline
\multicolumn{4}{|l|}{\textit{Imputation (ETTm1, Weather)}}  \\
\hline
-- 256 & 0.000 & 0.920 & 0.080 \\
-- 512 & 0.000 & 0.080 & 0.920 \\
-- 1024 & 0.800 & 0.120 & 0.080 \\
-- 2048 & 0.080 & 0.080 & 0.840 \\
\hline
\multicolumn{4}{|l|}{\textit{Outlier Detection (MSL, SMAP)}} \\
\hline
-- 256 & 0.000 & 0.000 & 1.000 \\
-- 512 & 0.750 & 0.000 & 0.250 \\
-- 1024 & 0.071 & 0.286 & 0.643 \\

 \hline 
\end{tabular}
\end{table}

The main insight from these results is the significant use of the MLP block in all cases. However, it remains relatively difficult to draw firm conclusions regarding block usage across different tasks or domains.

We believe part of the flexibility offered by \texttt{TimeBlocks} is its ability to automatically select the most suitable blocks, rather than relying on metadata-based rules. Introducing restrictions based on domain, data set, or task would shift the approach toward such rule-based selection.

\end{document}